\documentclass{article}

\usepackage{PRIMEarxiv}

\usepackage[utf8]{inputenc} 
\usepackage[T1]{fontenc}    
\usepackage{hyperref}       
\usepackage{url}            
\usepackage{booktabs}       
\usepackage{amsfonts}       
\usepackage{nicefrac}       
\usepackage{microtype}      
\usepackage{lipsum}
\usepackage{fancyhdr}       
\usepackage{graphicx}       
\graphicspath{{media/}}     
\usepackage{amsmath}
\usepackage{multirow}
\usepackage{subcaption}
\usepackage{mathtools}
\usepackage[ruled,vlined]{algorithm2e}
\usepackage{enumitem}

\title{Joint Demand Prediction for Multimodal Systems: A Multi-task Multi-relational Spatiotemporal Graph Neural Network Approach
}

\author{
  Yuebing Liang \\
  Department of Urban Planning and Design \\
  University of Hong Kong \\
  Hong Kong\\
  \texttt{yuebingliang@connect.hku.hk} \\
   \And
  Guan Huang \\
  Department of Urban Planning and Design \\
  University of Hong Kong \\
  Hong Kong\\
  \texttt{guanhuang@connect.hku.hk} \\
  \And
  Zhan Zhao\thanks{Corresponding Author} \\
  Department of Urban Planning and Design \\
  University of Hong Kong \\
  Hong Kong\\
  \texttt{zhanzhao@hku.hk} \\
}

\begin{document}
\maketitle

\begin{abstract}
Dynamic demand prediction is crucial for the efficient operation and management of urban transportation systems. Extensive research has been conducted on single-mode demand prediction, ignoring the fact that the demands for different transportation modes can be correlated with each other. Despite some recent efforts, existing approaches to multimodal demand prediction are generally not flexible enough to account for multiplex networks with diverse spatial units and heterogeneous spatiotemporal correlations across different modes. To tackle these issues, this study proposes a multi-relational spatiotemporal graph neural network (ST-MRGNN) for multimodal demand prediction. Specifically, the spatial dependencies across modes are encoded with multiple intra- and inter-modal relation graphs. A multi-relational graph neural network (MRGNN) is introduced to capture cross-mode heterogeneous spatial dependencies, consisting of generalized graph convolution networks to learn the message passing mechanisms within relation graphs and an attention-based aggregation module to summarize different relations. We further integrate MRGNNs with temporal gated convolution layers to jointly model heterogeneous spatiotemporal correlations. Extensive experiments are conducted using real-world subway and ride-hailing datasets from New York City, and the results verify the improved performance of our proposed approach over existing methods across modes. The improvement is particularly large for demand-sparse locations. Further analysis of the attention mechanisms of ST-MRGNN also demonstrates its good interpretability for understanding cross-mode interactions.
\end{abstract}


\section{Introduction} \label{sec:intro}

Urban transportation systems are generally multimodal in nature, consisting of several interconnected subsystems representing different modes of transportation, such as trains, buses, and cars. They are designed to meet diverse travel demand and provide urbanites with multiple travel options in cases of service disruptions. With growing urban population, limited capacities of transport infrastructure, and increasing concerns about urban resilience, it is more important than ever to plan, manage, and operate multimodal transportation systems in an integrated fashion. For example, ride-hailing services can be intelligently deployed to help people better access mass transit services, or replace certain transit trips when the transit system is overcrowded or delayed. Such multimodal operating strategies depend on the accurate and timely joint prediction of travel demand for different transportation modes, which is the focus of this study.

With the wide availability of mobility data and rapid advancement in computing technologies, short-term travel demand prediction has received much attention, though most studies have focused on demand prediction for a specific target mode. While earlier methods are based on various regression models (e.g., ARIMA), more recent efforts focus on deep learning-based approaches, particularly Graph Neural Networks (GNNs), due to their ability to extract complex spatiotemporal knowledge among large-scale mobility data \cite{li2017diffusion,geng2019spatiotemporal}. Despite the improved prediction performance, these methods still regard the target transportation mode as a closed system and ignore its potential interaction with other modes. In practice, there usually exist certain spatiotemporal correlations between different transportation modes through individual mode choices, passenger transfers (between modes), or trip chaining activities. For instance, the passenger flows of a subway station may affect the usage of ride-hailing service in the area since travelers may use ride-hailing as feeders to subway stations \cite{irawan2020compete}. Therefore, it is likely that the demand patterns of one mode can help us predict the future demand of a different mode. In addition, a deeper understanding of the intricate dependencies for multimodal travel demand can allow us to better formulate multimodal operating strategies to mitigate traffic congestion, improve user experience, and enhance system resilience. 

One major challenge for multimodal demand prediction is that different transportation modes have diverse spatial units: some are \textit{station-based} (e.g., subway), while others are \textit{stationless} (e.g., ride-hailing). For station-based modes, their operations are at the station level, and thus the demand prediction should match the same spatial granularity. For stationless modes, the operators usually define a number of service zones as the basic units of operations. To jointly model multimodal travel demand, recent works typically aggregate multimodal demand to a spatial grid \cite{ye2019co,wang2021learning} or other well-defined zone partitions \cite{ke2021joint}. Based on the same spatial structure, a similar model architecture can then be performed for different modes to learn shared spatiotemporal features \cite{wang2020multi}. These methods are generally less suitable for station-based modes, as a zone may contain 0 or multiple stations. Focusing on multimodal demand prediction for station-based public transit services, Li et al. \cite{li2021multi} developed a memory-augmented recurrent model for knowledge adaptation from a station-intensive mode to a station-sparse mode, though it only stores and shares mode-level temporal knowledge and is unable to leverage cross-mode spatial dependencies. To summarize, despite extensive research, there are still two important research gaps to be addressed:
\begin{itemize} [noitemsep]
    \item Most prior works focus on single-mode demand prediction and only consider intra-modal spatiotemporal correlations, ignoring its potential interaction with other modes. Inter-modal relationships exist because of complex travel behavior and can vary over space and time, making them difficult to model. 
    \item Existing approaches for multimodal demand prediction usually require aggregating multimodal demand data based on the same zone partition to enable shareable feature learning. These methods are unable to capture the cross-mode heterogeneous spatiotemporal correlations in general multimodal systems with multiplex networks and diverse spatial units. 
\end{itemize}

The overall objective of this study is to address the aforementioned issues by developing a multi-task multi-relational spatiotemporal graph neural network (ST-MRGNN) approach to multimodal demand prediction. Based on the proposed approach, the spatial dependencies across different spatial units (e.g., stations or zones) of different modes are encoded with multiple intra- and inter-modal relation graphs, with node-specific representations learned through a generalized graph convolution network and summarized via an attention-based aggregation module. Spatiotemporal blocks are also used to extract heterogeneous spatiotemporal relationships from the data. Empirical validation is done based on real-world multimodal datasets from New York City (NYC). The specific contributions of this study are summarized as follows: 
\begin{itemize} [noitemsep]
    \item We propose ST-MRGNN, a graph learning-based approach to demand prediction for multimodal systems. To the best of our knowledge, this is the first multimodal demand prediction model to account for heterogeneous spatiotemporal dependencies in multimodal systems with diverse spatial units.
    \item A multi-relational graph neural network (MRGNN) is developed to model spatial dependencies among diverse spatial units. Specifically, the cross-mode dependencies are encoded with multiple intra- and inter-modal relation graphs, the message passing mechanism within each relation graph is learned with a generalized graph neural network, and the information from different relations is summarized with an attention-based aggregation module. 
    \item We introduce the design of multi-relational spatiotemporal (ST-MR) blocks to jointly model spatiotemporal correlations between heterogeneous spatial units. Specifically, each ST-MR block captures mode-specific temporal patterns with gated convolution layers and fuses the heterogeneous temporal information in MRGNN layers.
    \item Extensive experiments are conducted based on real-world subway and ride-hailing datasets from NYC. The results show that the proposed approach outperforms existing methods across different modes, and the improvement is larger for demand-sparse locations.
\end{itemize}

\section{Literature Review} \label{sec:literature}

In this section, we review existing studies on short-term travel demand prediction under two categories, single-mode demand prediction and multimodal demand prediction. In addition, we present a short summary of recent works on heterogeneous graph embedding, which will be relevant to our proposed model. 

\subsection{Single-mode Demand Prediction} \label{literature:single}

Single-mode demand prediction is a well-studied problem. Early studies usually model travel demand as a time series based on various regression models, including ARIMA \cite{zhang2011seasonal}, local regression \cite{antoniou2013dynamic}, Kalman Filter \cite{lippi2013short} and Bayesian Inference \cite{feng2021multi}. For instance, Moreira-Matias et al. \cite{moreira2013predicting} developed an ensembled learning-based method for taxi demand prediction by combining ARIMA with time-varying Poisson models. Tong et al. \cite{tong2017simpler} proposed a linear regression model with massive features to forecast taxi demand. However, these methods are often less capable of capturing the nonlinearity and complex dependencies of large-scale mobility data, which can lead to relatively low prediction performance. 

Recently, deep learning has received growing interests because it provides a new solution to model the complex spatiotemporal relationships underlying travel demand. For example, Lv et al. \cite{lv2014traffic} applied a stacked autoencoder model for traffic flow prediction in freeway systems. Xu et al. \cite{xu2017real} devised recurrent neural networks (RNNs) to learn historical sequential patterns and make taxi demand predictions. Although these approaches demonstrate the effectiveness of deep neural networks for demand prediction, the spatial information is not explicitly considered. To tackle this issue, researchers employed convolutional neural networks (CNNs) to extract spatial correlations by dividing the study area into regular-shaped cells that resemble pixels in an image. Ke et al. \cite{ke2017short} proposed an on-demand ride service demand prediction model, using convolutional and Long Short-Term Memory (LSTM) networks to capture spatiotemporal dependencies simultaneously. Later, a multi-view learning framework was developed in \cite{yao2018deep} for taxi demand prediction, which incorporates CNNs and RNNs. Noursalehi et al. \cite{noursalehi2021dynamic} introduced a spatiotemporal model for the origin-destination (OD) demand prediction of urban rail systems with convolutional layers to capture the spatial dependencies within OD matrices. While CNNs work well for correlations in Euclidean space (e.g., a spatial grid), they are not applicable to non-Euclidean space, such as irregular service zones and unevenly distributed transit stations. In addition, CNN-based models only capture correlations between spatially adjacent neighborhoods and is insensitive to potential correlations between distant locations with similar functionalities.  

To address these problems, graph neural networks (GNNs) have been successfully adapted to various tasks in transportation. Li et al. \cite{li2017diffusion} first proposed diffusion graph convolutional networks (GCNs) and integrated it with recurrent layers for traffic forecasting. Yu et al. \cite{yu2017spatio} introduced a convolutional architecture for spatiotemporal prediction, with GCNs to capture spatial correlations and gated CNNs to capture temporal correlations. These methods assume that spatial dependencies are pre-determined by distance and neglect correlations among locations with similar functionalities or land use patterns. To uncover complex dependencies across mobility networks, Wu et al. \cite{wu2019graph} presented an adaptive learning technique, which learns an adaptive adjacency matrix to capture the hidden spatial dependency through node embedding. A multi-graph convolution network was developed in \cite{geng2019spatiotemporal}, which encodes the pair-wise correlations among locations using multiple graphs. Li et al. \cite{li2020graph} proposed a GCN-based model for public transit demand prediction using demand similarity to determine spatial dependencies. Liang et al. \cite{liang2021dynamic} provided a dynamic graph learning-based approach for traffic data imputation with a feed-forward network to learn dynamic spatial dependencies from real-time traffic condition.

Although the aforementioned approaches have achieved good performance in demand prediction, they are all developed for a single task (i.e., a single output variable). As travel demand may not be adequately described by a single variable, recent works have shifted more focus on multi-task demand prediction. Zhang et al. \cite{zhang2019short} proposed a LSTM-based model to jointly predict taxi pick-up and drop-off demand.  A multi-zone demand prediction model was presented in \cite{zhang2020taxi} using a convolutional multi-task learning network. Both Wang et al. \cite{wang2020multi} and Feng et al. \cite{feng2021multi} introduced a spatiotemporal architecture for co-prediction of zone-based and origin-destination-based demand values. In \cite{liu2021community}, the demand prediction for each zone is regarded as a distinct task, and an adaptive task grouping strategy was developed for community-aware multi-task demand prediction. However, these multi-task models are developed to jointly model multiple demand variables for the same transportation mode, and cannot be directly adapted for multimodal demand prediction. The latter requires the model to consider the inherent differences in spatial structures and demand patterns across different modes \cite{jiang2021graph}.

\subsection{Multimodal Demand Prediction} \label{literature:multi}

Limited attention has been paid to the problem of multimodal demand prediction. Ye et al. \cite{ye2019co} incorporated CNN and LSTM to jointly predict the pick-up and drop-off demand of taxis and shared bikes. In their research, the study area was divided into regular-shaped cells and the demand for taxis and shared bikes were aggregated to the same grid. Similarly, Wang et al. \cite{wang2021learning} proposed a convolutional recurrent network to co-predict travel demands for ride-hailing and bike sharing based on the same spatial grid. In \cite{ke2021joint}, a multi-graph learning-based approach was introduced to predict the zone-based ride-hailing demand for different service modes, i.e., solo-rides and shared-rides. Overall, these models all aggregate the demand data of different modes to the same zone partition. They are not suitable for general multimodal demand prediction, as multimodal systems often have heterogeneous network structures and diverse spatial units (e.g., subway stations vs ride-hailing zones), making the joint prediction more challenging. One exception is \cite{li2021multi}, who designed a memory-augmented recurrent architecture for knowledge adaptation from a station-intensive mode to a station-sparse mode. However, the memory-augmented network can only store and share mode-level temporal knowledge and is unable to utilize cross-mode spatial dependencies at the zone/station level. In this study, we aim to develop a multi-task learning framework for multimodal demand prediction that is capable of extracting complex spatiotemporal correlations among heterogeneous spatial units from different network structures. 

\subsection{Heterogeneous Graph Embedding} \label{literature:heterogeneous}

Although a plethora of GNN architectures have been developed for homogeneous graphs, which consist of one type of nodes and edges, they are not adaptable to heterogeneous graphs composed of different types of nodes and edges. To address this issue, some recent studies have proposed new techniques for heterogeneous graph embedding. Most of them focus on information networks, including knowledge graphs, social networks and recommendation systems \cite{wang2020survey}. For example, Schlichtkrull et al. \cite{schlichtkrull2018modeling} introduced relational GCNs to deal with knowledge graphs with different entities and applied them to link prediction and entity classification tasks. A composition-based GCN was proposed in \cite{vashishth2019composition} which jointly embeds both nodes and relations in a graph. To learn the importance between heterogeneous neighboring nodes, attention mechanisms have been employed in recent studies. For instance, Wang et al. \cite{wang2019heterogeneous} captured the heterogeneous semantic information across bibliographic networks with a hierarchical attention mechanism, including node-level and semantic-level attentions. Similarly, Zhang et al. \cite{zhang2020relational} proposed a GNN with two-level attentions for the knowledge graph completion task: the first-level attention encodes relation-level importance and the second-level attention encodes entity-level importance. However, all these studies are developed for information networks and related tasks, such as node classification, link prediction, and graph completion. Few studies have examined heterogeneous mobility networks, which is more challenging for two reasons. First, while information networks usually have pre-defined edges/relations, the correlations between nodes across mobility networks are more complex and not as well-defined. Second, unlike most information networks, mobility networks are dynamically changing over time as a result of congestion, service availability, human activities, and thus the temporal dimension should be considered alongside the spatial dimension. In this study, a multi-relational spatiotemporal graph neural network is introduced, which can effectively solve the above issues.

\section{Methodology} \label{sec:method}

In this section, we first define a few important concepts and formulate our problem. Next, we introduce a new spatiotemporal modeling framework for multimodal demand prediction, which incorporates multi-relational graph neural networks (MRGNNs) with temporal convolution networks (TCNs) to jointly model heterogeneous spatiotemporal correlations across multiple modes.

\subsection{Definitions and Problem Statement} \label{method:problem statement}

\textit{Definition 1 (Multimodal transportation System):} A multimodal transportation system $M$ is composed of $k$ $(k>1)$ transportation modes (e.g., subway, ride-hailing, etc.). Each transportation mode $m = 1, 2, ..., k$ has $N_m$ nodes (e.g., stations/zones) as the basic unit for passengers to travel from or to. For each node $i = 1, 2, ..., N_m$ at time step $t$, its outflow (or departure) and inflow (or arrival) demand can be represented as a 2-dimensional vector $x_{m, i}^t\in \mathbb{R}^2$. The demand of all the nodes from mode $m$ at time step $t$ is represented as $X_m^t=\{x_{m,0}^t, x_{m,1}^t…, x_{m,N_m}^t\},X_m^t \in \mathbb{R}^{N_m \times 2}$. Further, we use $X^t=\{X_m^t,\forall m\}$ to denote the demand of all the transportation modes at time step $t$. 

\textit{Definition 2 (Intra-modal Relation Graph):} To capture spatial correlations among stations/zones of the same mode, we define an intra-modal relation graph for each mode $m$, denoted as $G_m=(V_m, A_m)$, where $V_m$ is a set of nodes for mode $m$, $|V_m|=N_m$, and $A_m \in \mathbb{R}^{N_m \times N_m}$ is a weighted adjacency matrix representing the intra-modal dependencies between each pair of nodes in $V_m$. In the multimodal system $M$, a total of $k$ intra-modal relation graphs are defined, denoted as $G_{intra}=\{G_m, \forall m\}$.

\textit{Definition 3 (Inter-modal Relation Graph):} To capture the pairwise correlations among nodes from different modes, we further define an inter-modal relation graph between each mode pair $m, n = 1, 2, ..., k$ ($m \neq n$), which is represented as $G_{mn}=(V_m, V_n, A_{mn})$. $V_m$ and $V_n$ are nodes from mode $m$ and $n$ respectively and $A_{mn} \in \mathbb{R}^{N_m \times N_n}$ is a weighted matrix indicating the cross-mode dependencies between each node in $V_m$ and each node in $V_n$. Note that, unlike $A_m$, $A_{mn}$ is usually not a square matrix. A total of $\binom{k}{2}=k(k-1)/2$ inter-modal relation graphs are defined in the multimodal system $M$, denoted as $G_{inter}=\{G_{mn},\forall m, n\}$.

\textit{Problem (Multimodal Demand Prediction):} Given the historical observations of multimodal demand, and the intra- and inter-modal relation graphs, the multimodal demand prediction problem aims to jointly predict the demands for all transportation modes at the next time interval. 
Specifically, given the historical observations of two modes $m$ and $n$ denoted as $X_m^{t-T:t}, X_n^{t-T:t}$, the intra-model relation graphs $G_m, G_n$ and the inter-model relation graph $G_{mn}$
obtain a mapping function $F(\ast)$ to jointly predict the demands for all transportation modes at time interval $t+1$ denoted as $X^{t+1}$, given as:
\begin{equation}
X^{t+1}=F(X^{t-T:t}, G_{intra}, G_{inter}).
\end{equation}

While the problem is general, we will only focus on a bimodal system with subway and ride-hailing as a case study to demonstrate the proposed model in our experiments (see Section~\ref{sec:results}). To highlight the issue of heterogeneous spatial units, subway is chosen as an example of station-based modes, and ride-hailing an example of stationless modes. In this case, $k=2$ and there are 3 relation graphs defined to encode cross-mode correlations, including 2 intra-modal graphs and 1 inter-modal graph.

\subsection{Network Architecture}
\begin{figure}[ht!]
  \centering
  \includegraphics[width=\textwidth]{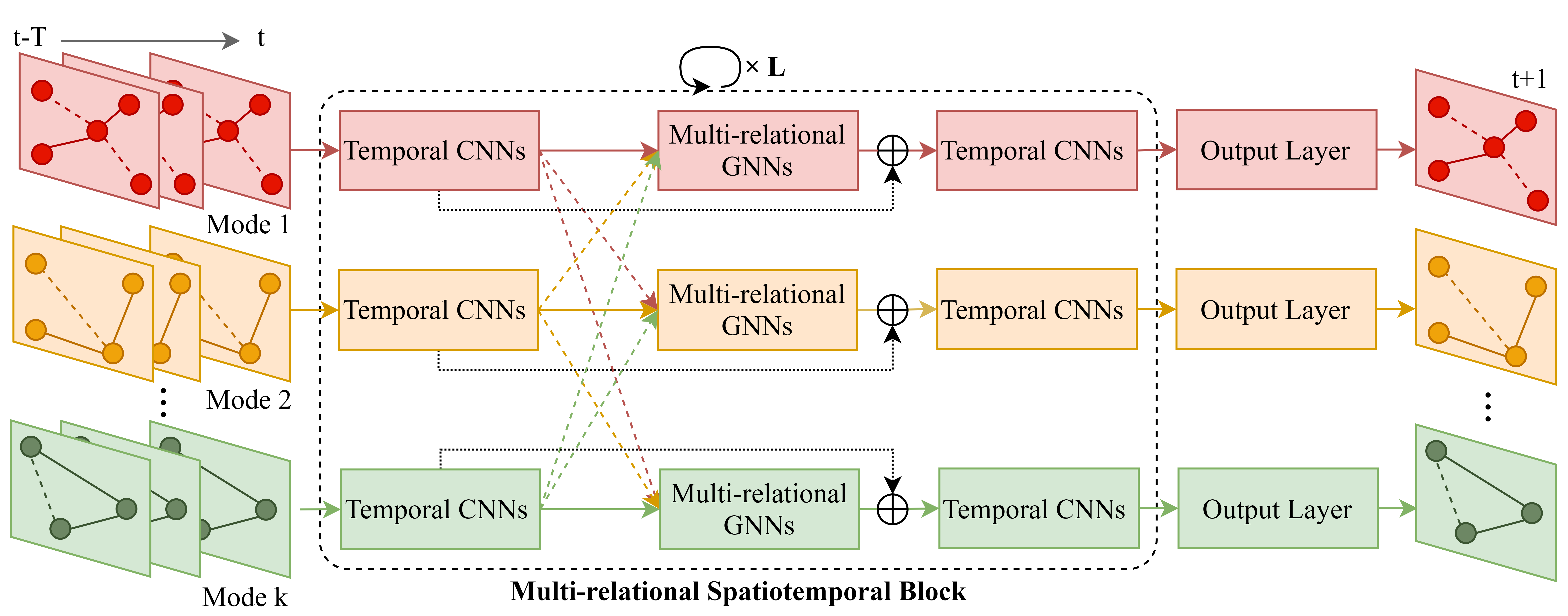}
  \caption{The architecture of ST-MRGNN ($\oplus$ denotes residual connections)}\label{fig:ST-MRGNN}
\end{figure}

Let us first introduce the overall framework of our proposed model. As shown in Figure~\ref{fig:ST-MRGNN}, ST-MRGNN is composed of $L$ multi-relational spatiotemporal blocks (ST-MR blocks) to uncover the heterogeneous spatiotemporal patterns across multiple modes, and an output layer for each mode to generate the final predictions. Each ST-MR block comprises several gated convolution layers(TCNs) to capture temporal features and a multi-relational graph neural network (MRGNN) to capture spatial features. Specifically, in each ST-MR block, a separate TCN layer is first applied to each mode to capture mode-specific temporal patterns. The extracted mode-specific features from TCN layers are then fused in a MRGNN layer to jointly model heterogeneous spatiotemporal dependencies for each mode. The MRGNN layer is followed by another mode-specific TCN layer because through experiments we find that it can help improve the prediction performance. It is potentially because that such a "sandwich" structure can facilitate fast spatial-state propagation between graph convolutions through TCN layers \cite{yu2017spatio}. The details of each module are introduced below.

\subsection{Multi-relational Graph Neural Network}
\begin{figure}[ht!]
  \centering
  \includegraphics[width=0.9\textwidth]{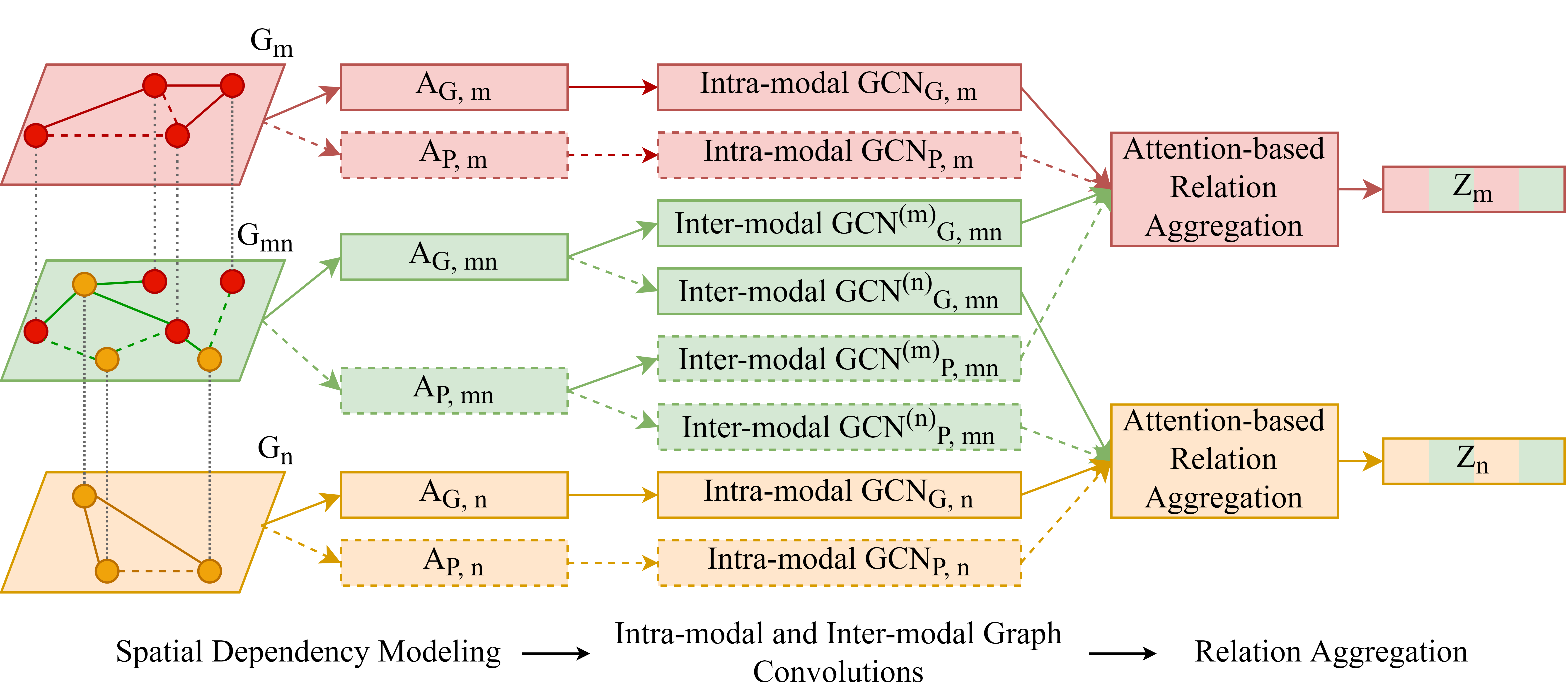}
  \caption{A bimodal framework for MRGNN}\label{fig:MRGNN}
\end{figure}

In this subsection, we introduce a novel graph neural network named MRGNN, which is capable of capturing heterogeneous spatial dependencies between nodes across multiple modes. To elaborate the key idea of MRGNN, we present the framework of MRGNN for a bimodal transportation system in Figure \ref{fig:MRGNN}. MRGNN consists of three major parts: (1) \textit{spatial dependency modeling}: two types of spatial dependencies are considered for each relation graph, namely geographical proximity and functional similarity; (2) \textit{intra- and inter-modal graph convolutions}: a generalized GCN is introduced to aggregate the node-level neighborhood information from intra- and inter-modal relation graphs; and (3) \textit{relation aggregation}: an attention-based aggregation module is designed to summarize the aggregated features from different relations. The three parts are described in details below.

\subsubsection{Spatial Dependency Modeling} \label{method:construction}
Previous research has shown that spatial dependencies exist between not only spatially adjacent locations, but also distant locations with similar functionalities and contextual environments \cite{geng2019spatiotemporal}. To capture both geographic and functional correlations among locations, we encode two types of spatial dependencies for each graph:

\textit{Geographical Proximity:} Locations that are geographically close to each other are likely to display strong correlations. We encode such geographical relationships among nodes using a distance-based adjacency matrix $A_G$. Formally, with the geographic center of each zone and station, we can compute $A_G$ as:
\begin{equation}
A_{G, ij} =
    \begin{cases}
      \exp(-(\frac{d_{ij}}{\sigma_d})^2) & d_{ij} \leq \kappa_d,\\
      0 & d_{ij} > \kappa_d,
    \end{cases}   
\end{equation}
where $A_{G, ij}$ is the weight of geographical proximity between nodes $i$ and $j$, $d_{ij}$ is the distance between $i$ and $j$, $\kappa_d$ is the distance threshold and $\sigma_d$ is the standard deviation of distances. 

\textit{Functional Similarity:} Locations that display similar demand patterns are also likely to share some common functionalities or other contextual features. To capture such semantic correlations, we construct an adjacency matrix $A_P$ for each graph. To deal with the varying demand of different modes, we first normalize the demand series of each mode and $A_P$ is given as:
\begin{equation}
A_{P, ij} = \frac{Corr(p_i, p_j)}{\sigma_{p,i} \sigma_{p,j}},
\end{equation}
where $A_{P, ij}$ indicates the weight of functional similarity between nodes $i$ and $j$, $p_i$ and $p_j$ are the normalized historical demand series of nodes $i$ and $j$, $Corr(\ast)$ calculates the correlation coefficient of two time series vectors, and $\sigma_{p,i}$ and $\sigma_{p,j}$ are the standard deviations of $p_i$ and $p_j$ respectively.

\begin{figure}[ht!]
  \centering
  \includegraphics[width=0.95\textwidth]{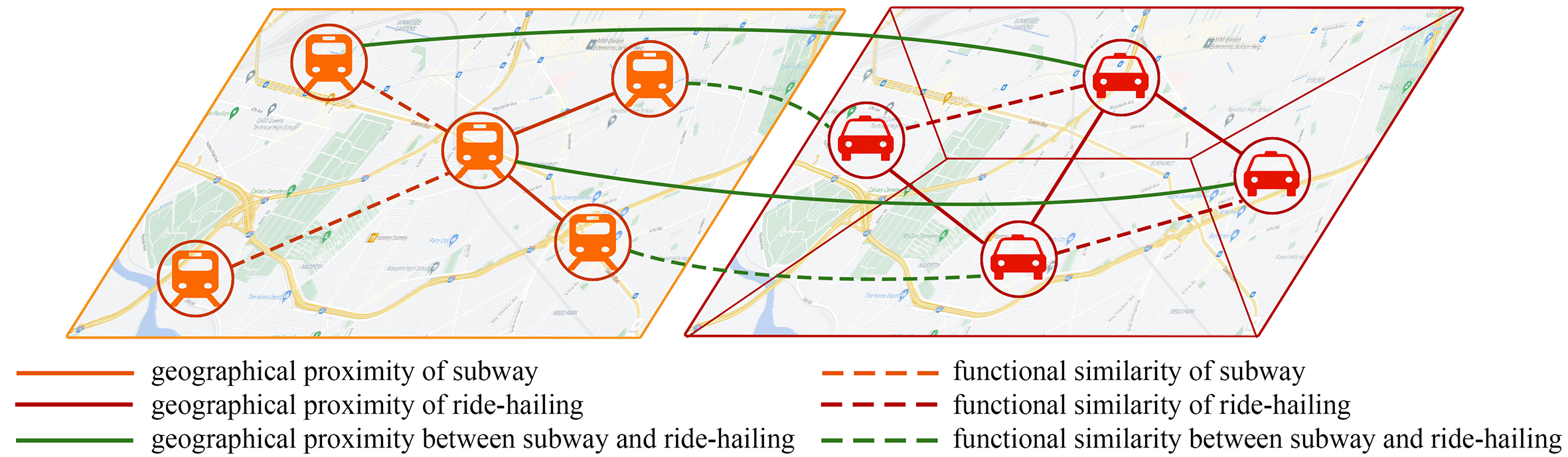}
  \caption{Modeling spatial dependencies for a multimodal system}\label{fig:graph construct}
\end{figure}

Generally, for each relation graph, we can model $u$ types of spatial dependencies. Therefore, in a multimodal system $M$, a total of $u \times (k + \binom{k}{2})$ relations can be encoded, including $u \times k $ intra-modal relations and $u \times \binom{k}{2}$ inter-modal relations. Note that this is the maximum number of relations to consider in the model, and not all of them are necessary depending on the system configurations and demand patterns. In our case where $k=2$ and $u=2$, there are 4 intra-modal relations and 2 inter-modal relations. Figure \ref{fig:graph construct} presents an example of a multimodal system with subway and ride-hailing. The solid and dashed lines are used to distinguish spatial dependencies based on geographic proximity or functional similarity. The line colors are used to distinguish cross-mode relations.

\subsubsection{Intra-modal and Inter-modal Graph Convolutions} \label{method:convolution}

Based on the multi-relational graphs introduced in Section \ref{method:construction}, each node is connected to heterogeneous nodes from multiple ($k$) modes via multiple ($u\times k$) relations. It is natural to aggregate the features of connected nodes from each relation using graph convolutions. However, most GCNs are developed for graph structures with a square adjacency matrix and cannot be applied to inter-modal relation graphs with heterogeneous nodes and a non-square adjacency matrix. To solve this issue, we introduce a generalized graph convolution network (GGCN), which is an extension of the standard GCN proposed in \cite{kipf2016semi}. Recall that the input of MRGNN is the mode-specific temporal features learned from TCN layers. Given an inter-modal relation graph $G_{mn} = (V_m, V_n, A_{mn})$ and the features extracted from TCN layers denoted as $H_m \in \mathbb{R}^{N_m \times c_{in}^s}$, $H_n \in \mathbb{R}^{N_n \times c_{in}^s}$, the graph convolution layer is then defined as:
\begin{gather}
Z_{mn}^{(m)} =  g({\widetilde{A}_{mn}^{(m)}} H_n W_{mn}^{(m)}+b_{mn}^{(m)}),\\
Z_{mn}^{(n)} =  g({\widetilde{A}_{mn}^{(n)}} H_m W_{mn}^{(n)}+b_{mn}^{(n)}),
\end{gather}
where $g(\ast)$ denotes a non-linear activation function ($ReLU$ in our case), and $Z_{mn}^{(m)} \in \mathbb{R}^{N_m \times c_{out}^s}$, $Z_{mn}^{(n)} \in \mathbb{R}^{N_n \times c_{out}^s}$ are the output aggregated features of mode $m$ on mode $n$ and mode $n$ on mode $m$ respectively, where $c_{in}^s$, $c_{out}^s$ are the input and output vector dimension of each node. $W_{mn}^{(m)}, W_{mn}^{(n)} \in \mathbb{R}^{c_{in}^s \times c_{out}^s}$ and $b_{mn}^{(m)}, b_{mn}^{(n)} \in \mathbb{R}^{c_{out}^s}$ are the learned model parameters. ${\widetilde{A}_{mn}^{(m)}} \in \mathbb{R}^{N_m \times N_n}$,$ {\widetilde{A}_{mn}^{(n)}} \in \mathbb{R}^{N_n \times N_m}$ are the normalized adjacency matrices constructed from $A_{mn}$ as:
\begin{gather}
{\widetilde{A}_{mn}^{(m)}} = \frac{A_{mn}}{rowsum(A_{mn})},\\
{\widetilde{A}_{mn}^{(n)}} = \frac{A_{mn}^T}{rowsum(A_{mn}^T)}.
\end{gather}

As introduced in Section \ref{method:construction}, each graph can be encoded with $u$ types of spatial dependencies. To process multiple dependencies on a graph at the same time, we extend GGCN to multi-dimensional tensors, given as:
\begin{gather}
\ddot{Z}_{mn}^{(m)} =  g({\ddot{A}_{mn}^{(m)}} H_n \otimes \ddot{W}_{mn}^{(m)}+b_{mn}^{(m)}),\\
\ddot{Z}_{mn}^{(n)} =  g({\ddot{A}_{mn}^{(n)}} H_m \otimes \ddot{W}_{mn}^{(n)}+b_{mn}^{(n)}),
\end{gather}
where $\otimes$ denotes the operation of batch matrix multiplication,  $\ddot{A}_{mn}^{(m)} \in \mathbb{R}^{u \times N_m \times N_n}$, $\ddot{A}_{mn}^{(n)} \in \mathbb{R}^{u \times N_n \times N_m}$ are stacked normalized adjacency matrices and $\ddot{W}_{mn}^{(m)}, \ddot{W}_{mn}^{(n)} \in \mathbb{R}^{u \times c_{in}^s \times c_{out}^s}$ are the parameter matrices. The outputs  $\ddot{Z}_{mn}^{(m)} \in \mathbb{R}^{u \times N_m \times c_{out}^s}, \ddot{Z}_{mn}^{(n)} \in \mathbb{R}^{u \times N_n \times c_{out}^s}$ are a set of cross-mode features between $m$ and $n$ from different spatial dependencies.  

An intra-modal relation graph can be regarded as a special case of inter-modal relation graphs when $m=n$. Therefore, given a mode $m$ and its intra-modal relation graph $G_m=(V_m, A_m)$, the correlations among nodes of mode $m$ are modeled as:
\begin{equation}
\ddot{Z}_{m} =  g({\ddot{A}_{m}} H_m \otimes \ddot{W}_{m}+b_{m}),
\end{equation}
where $\ddot{A}_{m}$ are stacked normalized adjacency matrices for mode $m$, and $\ddot{W}_{m}, b_m$ denote the model parameters.

Through the intra- and inter-modal graph convolutions, each node receives $u \times k$ aggregated feature vectors from its heterogeneous neighborhood nodes. Given a node $i$ in mode $m$, we denote the set of its aggregated features as:
\begin{equation}
{z}_{i}^{(m)} = \{{z}_{i}^{(m, n, r)}\}, \forall n \in \{1, 2, ..., k\}, r \in \{1,2,... u\},\\
\end{equation}
where ${z}_{i}^{(m, n, r)} \in \mathbb{R}^{c_{out}^s}$ is the aggregated features of nodes from mode $n$ to node $i$ in mode $m$ based on spatial dependency $r$. Note that ${z}_{i}^{(m, n, r)}$ denotes intra-modal relations when $m=n$ and inter-modal relations when $m \neq n$. 

\subsubsection{Relation Aggregation} \label{method:attention}
To summarize the learned features from different relations for each node, an intuitive operation is to simply add them. 
However, the contribution of different relations may vary for different nodes. For example, a ride-hailing zone that is close to a subway station may be easily influenced by inter-modal relations, while one that is distant from any subway station may rely more on intra-modal relations. To capture such variation, we design a relation-level attention module to learn the contribution of each relation to the target node. Given a node $i$ in mode $m$, the attention module is formulated as:
\begin{equation}
{a_i}^{(m)} = softmax(concat({z}_{i}^{(m)})W_{a}^{(m)} + b_{a}^{(m)}),
\end{equation}
where $concat(\ast)$ concatenates the learned features from all relations into a high-dimensional vector, $W_{a}^{(m)} \in \mathbb{R}^{(u \times k \times c_{out}^s) \times 1}, b_{a}^{(m)} \in \mathbb{R}$ are model parameters shared by all nodes in mode $m$. The resulting weight vector ${a_i}^{(m)} \in \mathbb{R}^{u \times k}$ is the learned attention weights with each element ${a_i}^{(m, n, r)}$ representing the contribution weight of nodes from mode $n$ to node $i$ in mode $m$ regarding spatial dependency $r$. Finally, node $i$ is represented as a weighted sum of the relation-specific features, given as:
\begin{equation}
{h}_{s, i}^{(m)} = \sum_{n=1}^{k}{\sum_{r=1}^{u}{{a}_{i}^{(m, n, r)}{z}_{i}^{(m, n, r)}}},
\end{equation}
where ${h}_{s, i}^{(m)} \in \mathbb{R}^{c_{out}^s}$ is the final output representations for node $i$ from the MRGNN layer. 

\subsection{Temporal Convolution Layer} \label{method:temporal}
We employ the temporal convolution network (TCN) proposed in \cite{yu2017spatio} for capturing temporal patterns of nodes in the multimodal system. Compared with RNN-based models that are widely used in time-series analysis, CNNs are advantageous in fast training time and simple structures. 
Given the input sequence of a node, the temporal convolution layer models the correlations between each time step and its $K_t$ neighborhoods using a 1-D causal convolution with a $K_t$-size kernel. Following \cite{yu2017spatio}, the convolution layer is conducted without padding, and therefore the output sequence length is shortened by $K_t-1$ each time. 

Previous research has shown that gating mechanisms are critical for temporal modeling in both RNNs and temporal CNNs \cite{wu2019graph}. To control the ratio of information that passes through layers, an output gate is incorporated in the convolution layer. Mathematically, given a node $i$ of mode $m$ and its input sequence $h_{in,i}^{(m)}$, the temporal gated convolution takes the form:
\begin{equation}
h_{c,i}^{(m)} = (W_{c,1}^{(m)} \star h_{in,i}^{(m)} + b_{c,1}^{(m)}) \odot \sigma(W_{c,2}^{(m)} \star h_{in,i}^{(m)} + b_{c,2}^{(m)}),
\end{equation}
where $h_{c,i}^{(m)}$ is the learned representations for node $i$ in mode $m$ from the TCN layer, $W_{c,1}^{(m)}$, $b_{c,1}^{(m)}$ are the model parameters for information learning, $W_{c,2}^{(m)}$, $b_{c,2}^{(m)}$ are the model parameters for computing the output gate, $\star$ is the convolution operation, $\odot$ is the element-wise product and $\sigma(\ast)$ represents the sigmoid function.

\subsection{Multi-relational Spatiotemporal Block} \label{method:block}

To integrate correlations from spatial and temporal domains, we incorporate MRGNN with TCN in a ST-MR block. Each ST-MR block is comprised of two TCN layers and a MRGNN layer in between. The input of the first ST-MR block is the historical multimodal demand series. In implementation, an equal MRGNN layer is applied to each time step in parallel. Similarly, the TCN layers are generalized to 3D tensors by employing the same convolution kernel to every node. To increase the speed of training, we employ a residual connection \cite{he2016deep} between TCN layers, given as:
\begin{equation}
\label{eq:st-block}
H_{\rho} = H_{c1} + H_{s},
\end{equation}
where $H_{\rho}$ is the input of the second TCN layer in a ST-MR block, and $H_{c1}$ and $H_{s}$ are the outputs of the first TCN layer and MRGNN layer, respectively. To stablize the model performance, layer normalizations \cite{ba2016layer} are implemented at the end of each ST-MR block.

Recall that the input length of historical demand sequence is $T$ and the sequence length is shortened by $K_t-1$ after each TCN layer. After stacking $L$ ST-MR blocks, the length of the output sequence from ST-MR blocks is shortened to $T-L \times (K_t-1) \times 2$. If the output sequence is still longer than one, an extra TCN layer is attached for each mode to downscale the outputs to a single time step. The output layer is a feed-forward network which maps the output signals of the ST-MR blocks to the prediction result of each mode.

\subsection{Training Strategy} \label{method:train}
The training objective of our proposed model is to minimize the difference between the real demand and the predicted one across all nodes from all modes. The loss function is defined as:
\begin{equation}
\label{eq:loss}
L(\theta) = \sum_{m=1}^{k}{\sum_{i=1}^{N_m}{\epsilon_m||\hat{x}_{m, i}^{t+1}-{x}_{m, i}^{t+1}||}},
\end{equation}
where $\hat{x}_{m, i}^{t+1}$, ${x}_{m, i}^{t+1}$ are the predicted and true demand values for node $i$ of mode $m$ at time step $t+1$ respectively, $\epsilon_m$ are the pre-determined weights to balance the loss of different modes and $\sum_{m \in M}{\epsilon_m}=1$. The training process of our proposed model is summarized in Alg.~\ref{alg:train}

\begin{algorithm}[ht!]
\SetAlgoLined
\SetKwInOut{KwIn}{Input}
\SetKwInOut{KwOut}{Output}
\KwIn{historical demand data $X^{t-T:t} =\{X_m^{t-T:t}, \forall m\}$, intra-modal relation graphs $G_{intra}$, inter-modal relation graphs $G_{inter}$, number of ST-MR blocks $L$, number of training epochs $E$}
 Initialize the parameters of ST-MRGNN\;
 \For{$e$ in $\{1, 2,... E\}$}{
  Initialize $H_{in, m}^{(1)}$ as $X_m^{t-T:t}$ for each mode $m\in \{1, 2,... k\}$\;
  \For{$l$ in $\{1, 2,... L\}$}{
    \For {$m \in \{1, 2,... k\}$}{
     Compute $H_{c1, m}^{(l)}$ from a mode-specific TCN layer with $H_{in, m}^{(l)}$ as input\;
    }
    $H_{c1}^{(l)} \longleftarrow \{H_{c1, m}^{(l)}, \forall{m \in \{1, 2,... k\}}\}$\;
    
    \For {$m \in \{1, 2,... k\}$}{
     Compute $H_{s, m}^{(l)}$ from a MRGNN layer given $H_{c1}^{(l)}, G_{intra}, G_{inter}$\;
     $H_{\rho, m}^{(l)} \longleftarrow H_{c1, m}^{(l)} + H_{s, m}^{(l)}$\;
     Compute $H_{c2, m}^{(l)}$ from a TCN layer with $H_{\rho, m}^{(l)}$ as input\;
     $H_{out,m}^{(l)} \longleftarrow LayerNormalization(H_{c2, m}^{(l)})$\;
     $H_{in,m}^{(l+1)} \longleftarrow H_{out,m}^{(l)}$\;
    }
  }
  \For {$m \in \{1, 2,... k\}$}{
     Generate predictions $\hat{X}_m^{t+1}$ from $H_{out,m}^{(L)}$ using a mode-specific output layer\;
  }
  Calculate the loss function (Eq. \ref{eq:loss}) from prediction results $\{\hat{X}_m^{t+1}, \forall{m \in \{1, 2,... k\}}\}$\;
  Update the parameters through error back propagation\;
 }
 \caption{The training process of ST-MRGNN}\label{alg:train}
\end{algorithm}

\section{Experiments} \label{sec:results}

\subsection{Data Description} \label{res:data}

To verify the effectiveness of our proposed model, we conduct experiments on real-world multimodal datasets from New York City (NYC). Specifically, we use subway as an example of station-based modes and ride-hailing an example of stationless modes. We choose these two datasets because previous research has shown that subway and ride-hailing are both important components of urban transportation systems and exhibit strong spatiotemporal correlations especially in metropolitan areas \cite{irawan2020compete}. In this study, we use Manhattan as the research area and both data are collected from March 1st, 2018 to August 31th, 2018.

\textit{NYC Subway}\footnote{\href{https://toddwschneider.com/dashboards/nyc-subway-turnstiles}{https://toddwschneider.com/dashboards/nyc-subway-turnstiles}}: The data provides the turnstile usage counts of NYC subway stations every 4 hours. There are 2,289 turnstiles used as entry/exit registers at 136 subway stations in our study area. During the study period, there are about 2.4 million entry/exist counts every day on average. The original data consists of the following information: station ID, turnstile ID, record time, entries and exists counts, etc.

\textit{NYC Ride-hailing}\footnote{\href{https://www1.nyc.gov/site/tlc/about/tlc-trip-record-data.page}{https://www1.nyc.gov/site/tlc/about/tlc-trip-record-data.page}}: The for-hire vehicle data is made available by NYC Taxi \& Limousine Commission (TLC) and built on data from ride-hailing companies such as Uber and Lyft. It consists of 43 million trip records during the study period with 234 thousand trips per day on average. For each trip, the data provides the following information: pick-up time, pick-up zone, drop-off time, drop-off zone, etc. The zones are pre-determined by TLC. There are 63 TLC zones in total in our study area.

\begin{figure}
    \centering
    \includegraphics[width=0.85\textwidth]{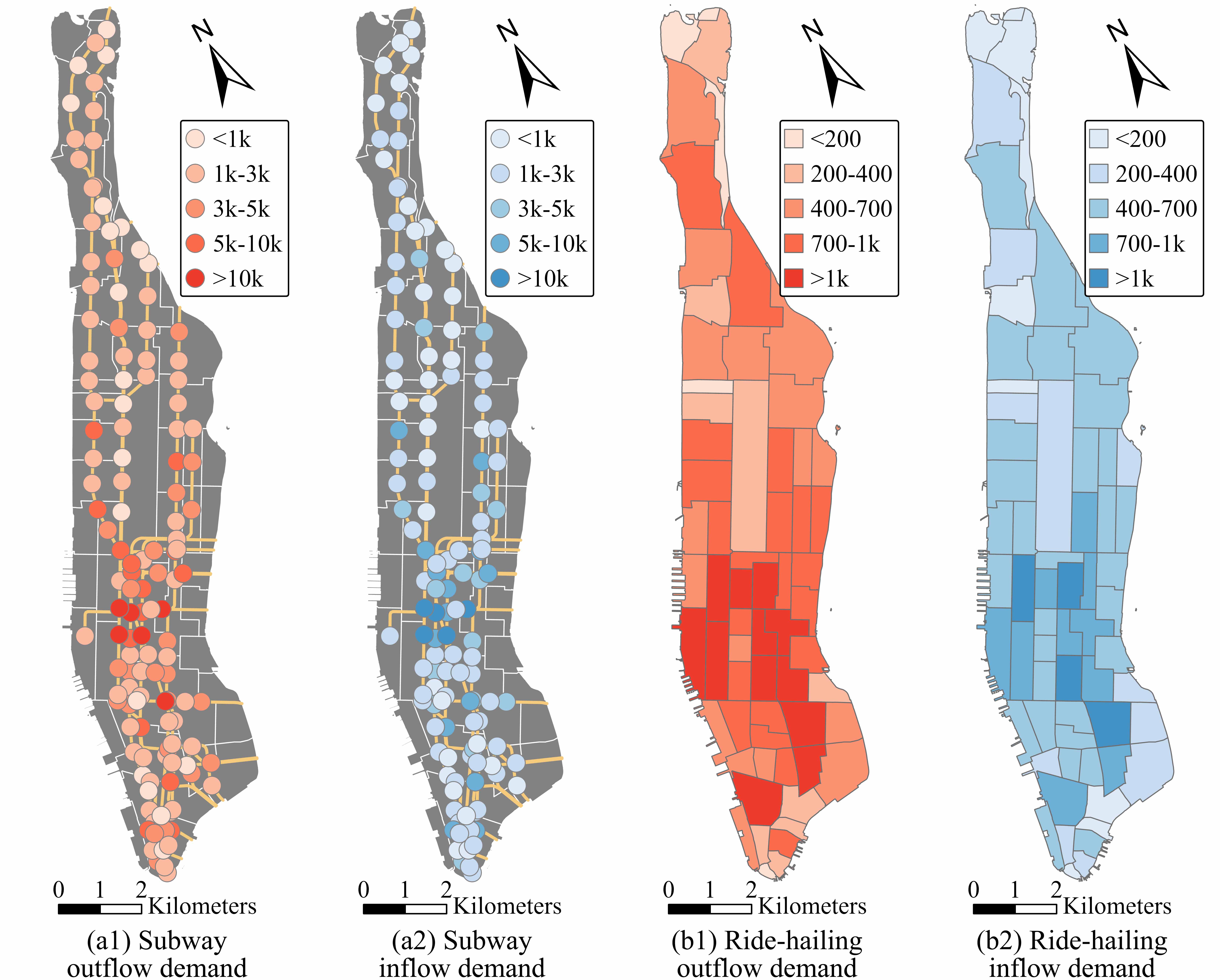}
    \caption{Spatial distribution of multimodal travel demand in Manhattan}
    \label{fig:od_s}
\end{figure}

\begin{figure}
    \centering
    \includegraphics[width=\textwidth]{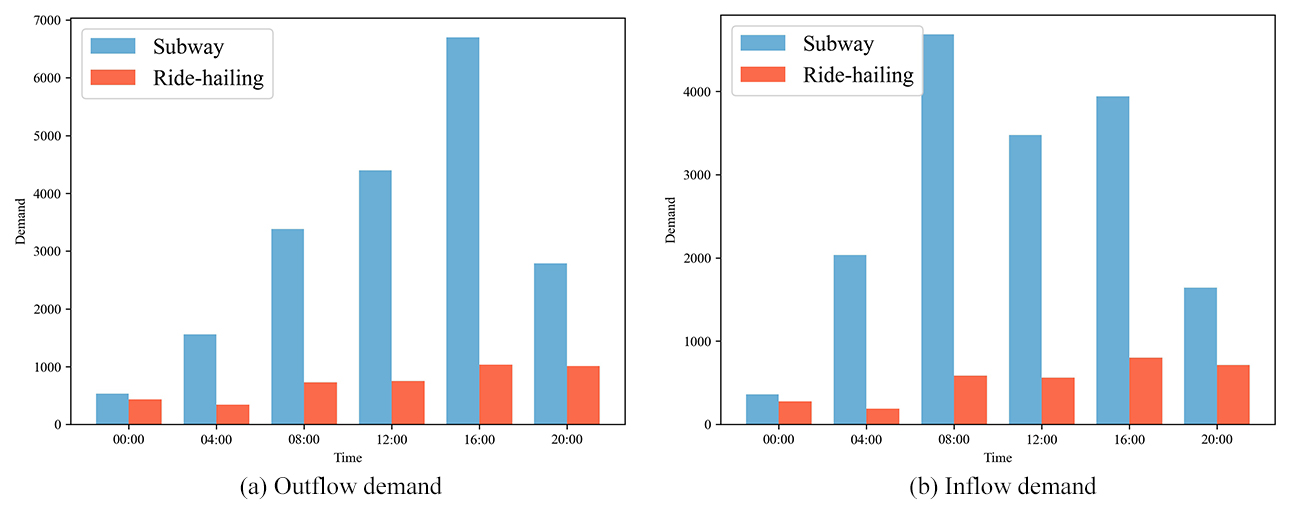}
    \caption{Temporal pattern of multimodal travel demand in Manhattan}
    \label{fig:od_t}
\end{figure}

The spatial distribution and temporal pattern of the multimodal data in NYC are presented in Figures~\ref{fig:od_s} and \ref{fig:od_t}. It can be clearly seen from Figure~\ref{fig:od_s} that the spatial units of different modes are heterogeneous: subway is station-based while ride-hailing is zone-based. The demand for subway and ride-hailing exhibit correlated spatial distributions: the stations/zones with the most intensive demand are mainly in Midtown Manhattan, followed by Downtown, and finally Uptown. Figure~\ref{fig:od_t} compares the average demand density of each station/zone every four hours for different modes. Obviously, subway has much larger travel demand than ride-hailing. The demand patterns of them are also temporally correlated, though the subway demand shows a more pronounced commuting pattern. Since many New Yorkers reside in outer boroughs and commute to Manhattan for work, we can see a clear afternoon peak for subway outflows and a higher morning peak for subway inflows.

\subsection{Baseline Models} \label{res:baseline}
As introduced in Section \ref{literature:multi}, most existing multimodal demand prediction approaches require demand aggregation on the same zone partition and are not suitable for multimodal systems with diverse spatial units. The only exception is \cite{li2021multi}, who proposed a memory-augmented recurrent model for knowledge adaptation from a station-intensive mode to a station-sparse mode. Since the source code is not available, we were unable to replicate the same model, and our own implementation does not perform as well as other baselines. Therefore, we only present the baseline results of single-mode prediction methods as listed below:

\begin{itemize} [noitemsep]
    \item \textbf{Historical Average (HA)}: a statistical method which predicts future demand with average values of previous time steps.
    \item \textbf{Linear Regression (LR)}: a regression approach to modeling the relationship between future demand and historical time series.
    \item \textbf{eXtreme Gradient Boosting (XGBoost)}\cite{chen2015xgboost}: a representative machine learning method based on gradient boosting decision trees.
    \item \textbf{Long-short Term Memory (LSTM)}\cite{xu2017real}: a deep learning model that incorporates LSTM with feed-forward networks for time series modeling.
    \item \textbf{Spatiotemporal Graph Convolution Network (STGCN)} \cite{yu2017spatio}: a GCN-based method that models spatial correlations with spectral-based GCNs and temporal dependencies with temporal convolution layers.
    \item \textbf{Multi-graph Convolution Network (MGCN)} \cite{geng2019spatiotemporal}: a multi-graph convolution network that captures multiple types of spatial correlations with multiple GCNs and temporal dependencies with contextual RNNs.
    \item \textbf{Graph WaveNet} \cite{wu2019graph}: a spatiotemporal graph learning appproach using a self-learned adjacency matrix to capture complex spatial dependencies through node embedding.
\end{itemize}

\subsection{Experiment Settings} \label{res:settings}
To align different datasets, we aggregate the multimodal demand data into 4-hour intervals and apply min-max normalization to each mode. We choose 4-hour intervals because the NYC subway demand data is also aggregated every 4 hours. The historical time step $T$ is set as 6 (i.e., $6 \times 4 = 24$ hours).  For all the deep learning models, we use data from the first 60\% time steps for training, the following 20\% for validation and the last 20\% for test. The number of training epochs $E$ is set as 500 and we use early stopping on the validation set to prevent overfitting. The models are trained using Adam Optimizer with a learning rate of 0.002, a batch size of 32, a dropout rate of 0.3 and L2 regularization with a weight decay equal to 1e-5.

Through extensive experiments, we determine the hyperparameters of our proposed model as follows: the number of ST-MR blocks $L=2$, the width of the temporal convolution kernel $K_t=2$, the input and output dimensions of TCN layers $c_{in}^{t}=16, c_{out}^{t}=64$, the input and output dimensions of MRGNN layers $c_{in}^{s}=64, c_{out}^{s}=16$ and the hidden dimension of output layers $c_h=128$. The hyperparameters for loss balancing $\epsilon_m$ are set as 0.5, 0.5 for subway and ride-hailing respectively. For baseline models, we implement XGBoost and LSTM ourselves and fine-tune their hyperparameters based on the validation set. For STGCN and Graph WaveNet, we use the open source codes provided by the original authors with their default parameter settings. For MGCN, the original performance is poor, and thus we add an extra GCN layer for each correlation through experiments. The models are evaluated with three commonly used metrics computed on the test set: Root Mean Square Error (RMSE), Mean Absolute Error (MAE) and Coefficient of Determination ($R^2$).

\section{Results}
\subsection{Comparison of Model Performance} \label{res:performance}
Table~\ref{table:baseline_model} summarizes the demand prediction performance of different models for NYC subway and ride-hailing. Each model is run 10 times and the average values are reported. For all models, the RMSE of subway is much larger than ride-hailing. This is because there are much more subway trips than ride-hailing in NYC. The subway is also associated with higher $R^2$, indicating that the subway has more regular and predictable demand patterns than ride-hailing. Compared with baseline models, our proposed model ST-MRGNN achieves the best performance on all evaluation metrics across modes, suggesting that both modes can benefit from the knowledge of demand patterns of other modes. 

Among baseline models, HA and LR perform worse than the other methods, indicating the limitation of classical statistical models in extracting nonlinear spatiotemporal correlations. XGBoost shows poor performance for the prediction of ride-hailing demand, but achieves comparable performance with LSTM for subway, implying the effectiveness of ensemble methods in some cases. Among deep learning models, the poor performance of LSTM indicates the necessity to consider spatial dependencies for demand prediction. MGCN outperforms STGCN for subway and ride-hailing, demonstrating the value of considering multiple types of spatial dependencies. Benefiting from the adaptive adjacency matrix, Graph WaveNet achieves competitive results for both modes. Our proposed model is shown to achieve significantly better prediction performance than Graph WaveNet, with 5.04\% and 11.49\% improvements in RMSE and MAE for subway, and 12.98\% and 15.05\% improvements in RMSE and MAE for ride-hailing. While most existing models only consider intra-modal dependencies, ST-MRGNN can effectively account for complex spatiotemporal relationships between modes. Compared with RMSE and MAE, the relative improvement in $R^2$ for our model is relatively small, but still significant.

\begin{figure}[ht!]
  \centering
  \includegraphics[width=\textwidth]{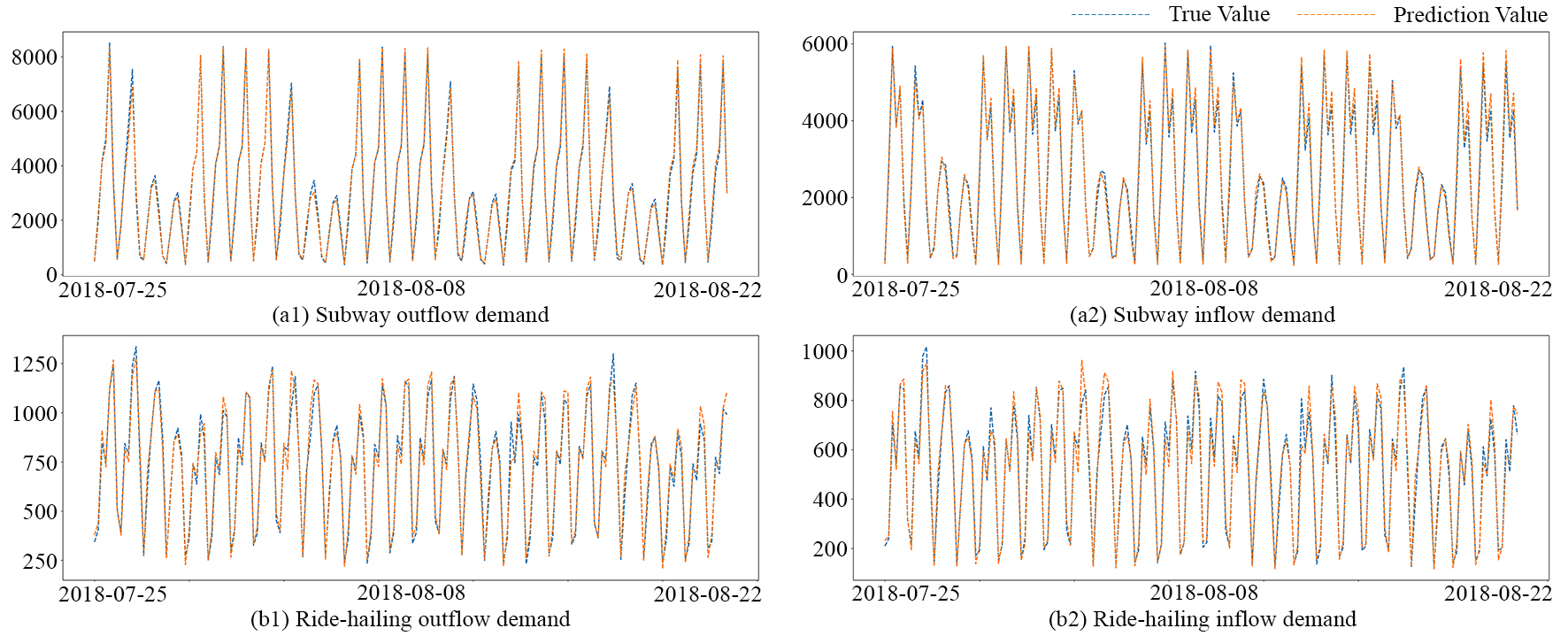}
  \caption{The average prediction and true values during 2018-07-25 to 2018-08-22}\label{fig:predict_true}
\end{figure}

To intuitively illustrate the demand prediction results, we compare the average predicted and true values in the first four weeks of our test set. As shown in Figure~\ref{fig:predict_true}, the predicted curves (in orange) can accurately trace the actual curves (in blue). Specifically, the demands for subway exhibit strong temporal periodicity while the ride-hailing demand shows more irregular fluctuations, likely because that people use ride-hailing not only for commuting but also for various purposes (e.g., leisure) that have less spatiotemporal regularities. As a result, the ride-hailing demand is harder to predict, as evidenced by the lower $R^2$. At the same time, our proposed model can achieve more improved prediction performance for ride-hailing, as some of its demand irregularities may be explained by multi-relational intra- and inter-modal dependencies. 

\begin{table}[ht!]
  \centering \footnotesize
  \caption{Performance comparison of different models on NYC Dataset}
    \begin{tabular}{cccccccccc}
    \toprule
    \multirow{2}{*}{Models} & \multicolumn{3}{c}{Subway} & \multicolumn{3}{c}{Ride-hailing} \\
    & RMSE & MAE & $R^2$ & RMSE & MAE & $R^2$\\
    \midrule
    HA & 3243.228 & 1668.378 & 0.499 & 372.972 & 263.645 & 0.458  \\
    LR & 1793.461 & 729.990 & 0.844 & 195.965 & 117.721 & 0.848 \\
    XGBoost & 1390.444 & 638.448 & 0.907 & 171.203 & 101.337 & 0.885 \\
    LSTM &  1222.559 & 707.432 & 0.930 & 146.457 & 98.055 & 0.917 \\
    STGCN & 984.745 & 569.768 & 0.966 & 111.052 & 76.036 & 0.956 \\
    MGCN & 888.024 & 491.382 & 0.963 & 106.468 & 71.552 & 0.957 \\
    Graph WaveNet & 683.558 & 389.887 & 0.978 & 101.931 & 69.710 & 0.960 \\
    ST-MRGNN & \underline{\textit{649.084}} & \underline{\textit{345.108}} & \underline{\textit{0.981}} & \underline{\textit{88.704}} & \underline{\textit{59.217}} & \underline{\textit{0.971}} \\
    \bottomrule
    \end{tabular}%
  \label{table:baseline_model}%
\end{table}%

\begin{figure}[ht!]
  \centering
  \includegraphics[width=0.85\textwidth]{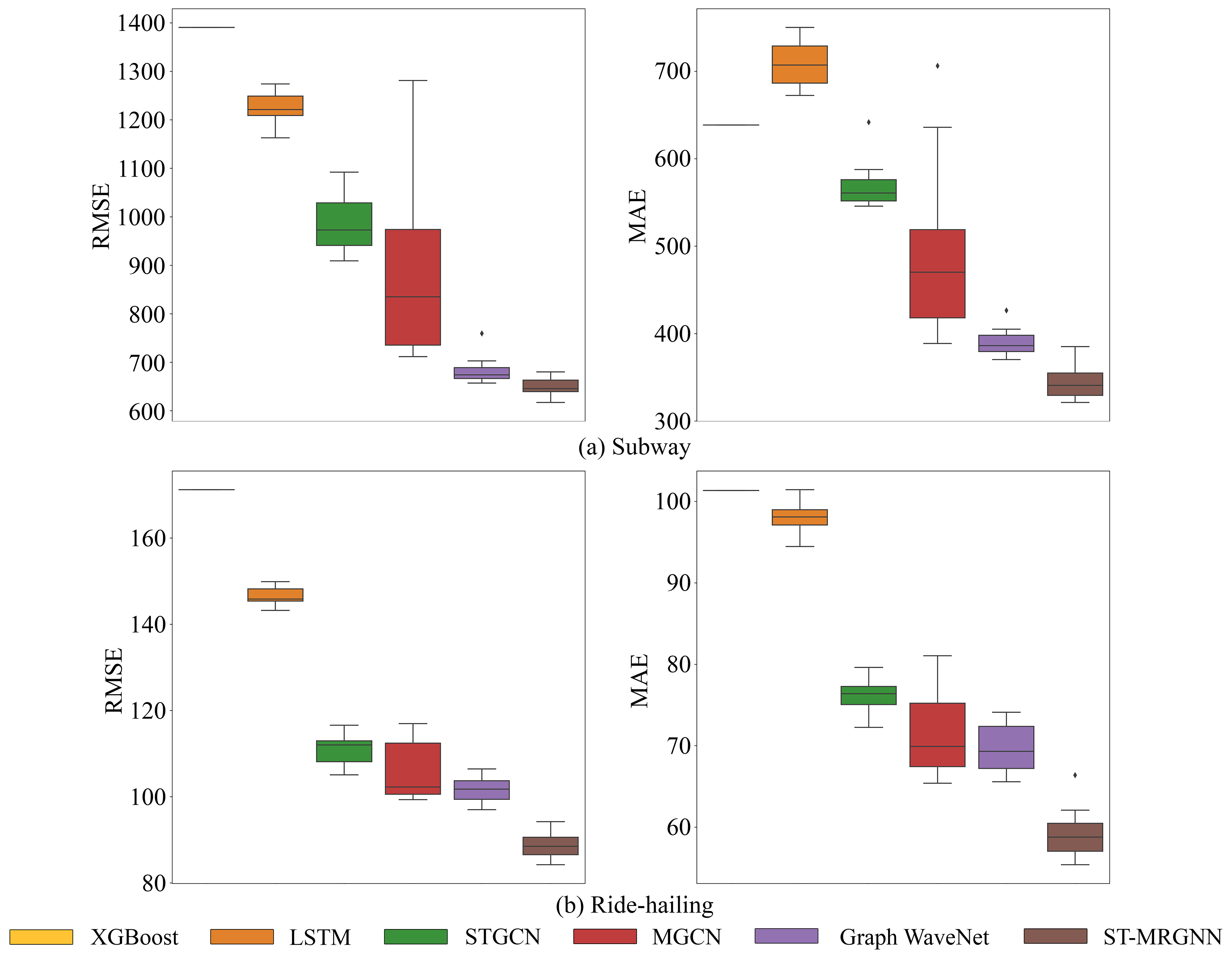}
  \caption{Comparison of model stability}\label{fig:boxplot}
\end{figure}

Figure~\ref{fig:boxplot} compares the stability of a few selected models based on 10 runs. It is found that our method has the best performance in most cases with relatively small performance variability. For ride-hailing demand prediction, our proposed model outperforms Graph WaveNet significantly in almost all experiments. For subway demand prediction, ST-MRGNN achieves the lowest RMSE in more than 75\% experiments. Among baseline models, MGCN exhibits relatively poor model stability. This is potentially because it employs a model structure in which different types of correlations are modeled with separate GCN layers in parallel. Unlike MGCN, ST-MRGNN and Graph WaveNet both use stacked spatiotemporal blocks with GCN layers to encode spatial correlations, which enhances the mutual connection of different blocks and improves the model stability.

Furthermore, we compare the computation cost of STGCN, MGCN, Graph WaveNet and ST-MRGNN as displayed in Table~\ref{table:computation}. It is shown that STGCN is most efficient among the four models due to its simple architecture. ST-MRGNN runs two times faster than Graph WaveNet and similar to MGCN for training. For inference, we measure the total time cost of each model on the test set. ST-MRGNN is much faster than both Graph WaveNet and MGCN. This is likely because our model evaluates two modes at the same time while the other models evaluate different modes separately. These results show that our proposed model can achieve more accurate prediction performance with relatively low computation cost.

\begin{table}[ht!]
  \centering \footnotesize
  \caption{Comparison of computation efficiency}
  \resizebox{\linewidth}{!}{%
    \begin{tabular}{ccccccccc}
    \toprule
    \multirow{2}{*}{Model} & \multicolumn{4}{c}{Training Time (s/epoch)}&\multicolumn{4}{c}{Inference Time(s)}\\
     & STGCN & MGCN & Graph WaveNet & ST-MRGNN& STGCN & MGCN & Graph WaveNet & ST-MRGNN\\
    \midrule
    Subway & 0.946 & 1.407 & 3.912 & - & 0.071& 0.459 & 0.409 & -\\
    Ride-hailing & 1.038 & 1.359 & 1.890 & - & 0.044& 0.293 & 0.206 & -\\
    Total & 1.984 & 2.766 & 5.802 & 2.710 & 0.115 & 0.752 & 0.615 & 0.154\\
    \bottomrule
    \end{tabular}%
    }
  \label{table:computation}%
\end{table}%

\subsection{Comparison of Model Variants} \label{res:ablation}

To quantify the contribution of different components in ST-MRGNN, we further implement four simplified versions of ST-MRGNN for ablation tests. Each variant model drops a component of ST-MRGNN as listed below:
\begin{itemize} [noitemsep]
    \item \textbf{Inter-modal relation graphs ($InterGraph$)}: To compare the model performance with or without inter-modal relation graphs, we implement a single-mode version of ST-MRGNN, which models each mode separately with only intra-modal relation graphs.
    \item \textbf{Geographical proximity ($A_G$)}: ST-MRGNN encodes two types of spatial dependencies: geographical proximity and functional similarity. With geographical proximity ablated, the variant model only uses functional similarity for encoding spatial dependencies.
    \item \textbf{Functional similarity ($A_P$)}: In this variant, with the dependency of functional similarity ablated, the spatial correlations among nodes are only encoded based on geographical proximity.
    \item \textbf{Attention-based aggregation module ($AttnAgg$)}: Without the attention module, the variant model aggregates the outputs of different relations by simply adding them. 
\end{itemize}

Table~\ref{table:ablation} displays the performance comparison of ST-MRGNN and its variant models with different model components ablated. We find that the cross-mode dependencies, captured by the inter-modal relation graphs in our model, are crucial for the performance improvement for both datasets. Without cross-mode dependencies, the RMSE for subway and ride-hailing increases by 8.34\% and 4.85\% respectively. This shows that the inter-modal relations from geographically nearby or functionally similar subway stations and ride-hailing zones can indeed help with the demand prediction for each other. It is also worth noting that, for ride-hailing demand prediction, the single-mode version of ST-MRGNN already outperforms the baseline models listed in Table~\ref{table:baseline_model}. This indicates that for a single stationless mode (i.e., ride-hailing in our case), our proposed model structure, which consists of stacked spatiotemporal blocks with multi-relation graphs, already shows superiority in extracting intra-modal spatiotemporal correlations. Removing the spatial dependency of either geographic proximity or functional similarity leads to worse performance for both modes, validating the importance of considering both types of dependencies. Between the two, the functional similarity has a greater impact on both subway and ride-hailing demand prediction, suggesting that functionally similar stations/zones contribute more to the demand prediction of different transportation modes. With the attention module ablated, the RMSE increases by 3.71\% for subway and slightly for ride-hailing, demonstrating that the attention mechanism can help find the optimal combination of different relations across modes, especially for subway. 

\begin{table}[ht!]
  \centering \footnotesize
  \caption{Ablation analysis of different ST-MRGNN components}
    \begin{tabular}{ccccccc}
    \toprule
    \multirow{2}{*}{Models} & \multicolumn{3}{c}{Subway} & \multicolumn{3}{c}{Ride-hailing} \\
     & RMSE & MAE & $R^2$ & RMSE & MAE & $R^2$\\
    \midrule
    ST-MRGNN & \underline{\textit{649.084}} & \underline{\textit{345.108}} & \underline{\textit{0.981}} & \underline{\textit{88.704}} & \underline{\textit{59.217}} & \underline{\textit{0.971}} \\
    - $InterGraph$ & 713.227 & 383.177 & 0.977 & 93.006 & 62.830 & 0.968  \\
    - $A_G$ & 656.621 & 349.254 & 0.981 & 91.821 & 61.900 & 0.970 \\
    - $A_P$ &  670.964& 354.769 & 0.980 & 92.327 & 62.109 & 0.970 \\
    - $AttnAgg$ & 673.168 & 366.883 & 0.980 & 89.447 & 59.384 & 0.971\\
    \bottomrule
    \end{tabular}%
  \label{table:ablation}%
\end{table}%

\subsection{Comparison of Multimodal and Single-mode ST-MRGNN} \label{res:multi_vs_single}
\begin{figure}[ht!]
    \centering
    \includegraphics[width=0.4\textwidth]{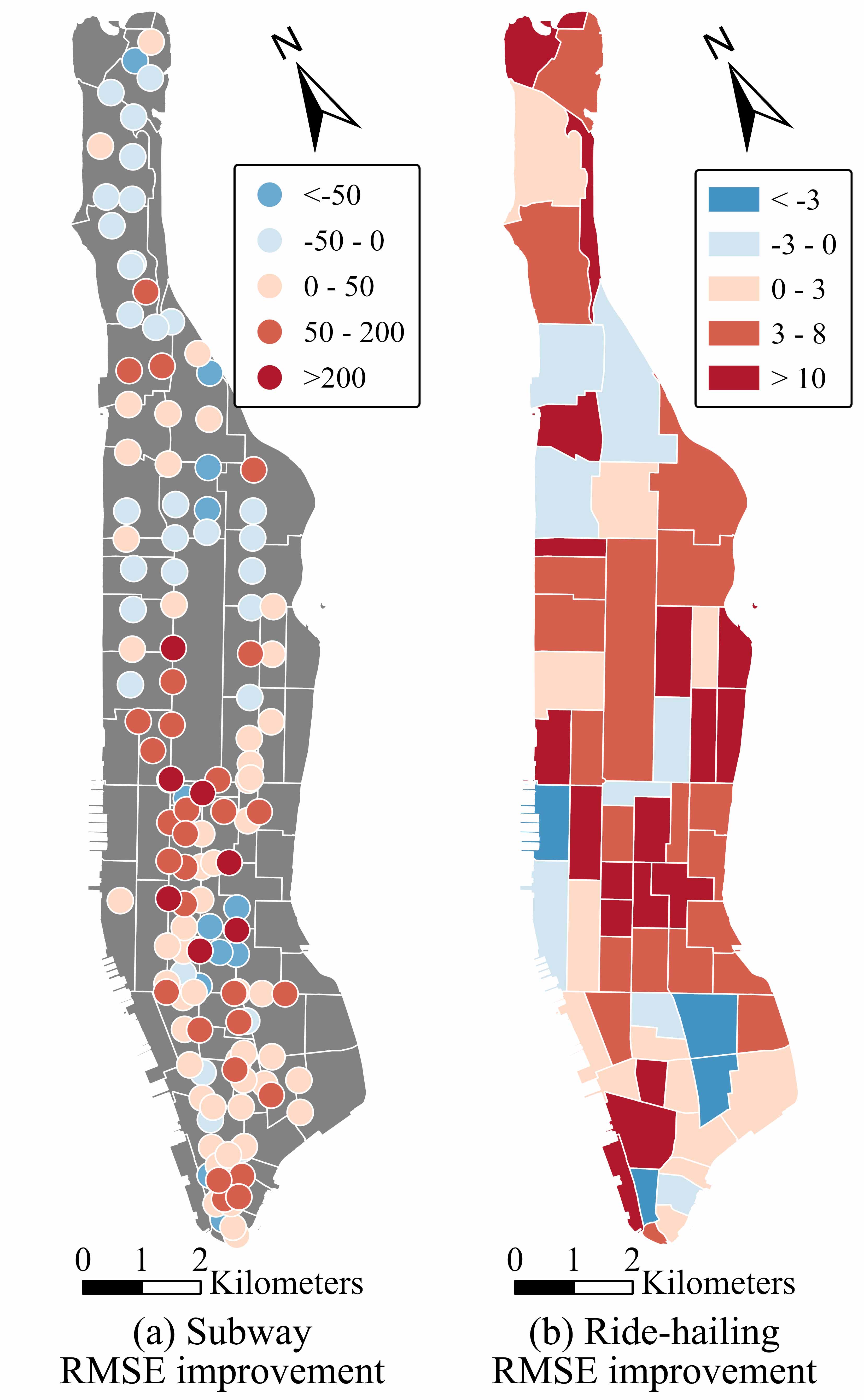}
    \caption{The distribution of RMSE improvement of ST-MRGNN against S-ST-MRGNN}
    \label{fig:diff}
\end{figure}

To further investigate the effect of inter-modal relations on the model performance, we compare ST-MRGNN with the single-mode version of ST-MRGNN, named S-ST-MRGNN, and examine the differences across stations/zones. As introduced in Section~\ref{res:ablation}, S-ST-MRGNN has a similar network architecture as ST-MRGNN. The only difference is that S-ST-MRGNN models each mode separately without inter-modal relations. We first categorize the stations/zones of each mode into different groups according to their demand density levels. Specifically, the subway stations or ride-hailing zones are sorted according to their average demand levels, and then the top 1/3 are classified as demand-intensive (DI) locations and the bottom 1/3 as demand-sparse (DS) locations. Table~\ref{table:demand_intensive_sparse} summarizes the demand prediction performance of S-ST-MRGNN and ST-MRGNN for DI and DS locations. Obviously, for both modes, the RMSE of DI locations is much larger than DS ones, which is reasonable as the former have higher demand density. 
DI locations are also associated with relatively higher $R^2$. Unlike RMSE, $R^2$ is normalized and scale-free. Therefore, the results indicate that the subway stations or ride-hailing zones with sparser demand are less predictable. They are likely associated with more uncertainty in demand patterns, and thus more challenging to estimate with high confidence. Compared with S-ST-MRGNN, ST-MRGNN achieves better performance across different stations/zones, particularly for DS locations. This implies that DS locations can potentially benefit more from the knowledge of demand patterns for other modes. 
Between the two modes, using S-ST-MRGNN, the $R^2$ of DS subway stations is much lower than DS ride-hailing zones, which is potentially because the demand density of different subway stations varies more drastically than ride-hailing zones (see Figure~\ref{fig:od_s}). Meanwhile, using ST-MRGNN, the relative improvement in $R^2$ for DS subway stations is much higher than DS ride-hailing zones. Figure~\ref{fig:diff} further illustrates the RMSE improvement of ST-MRGNN against S-ST-MRGNN for each station/zone. For subway, ST-MRGNN outperforms S-ST-MRGNN in most stations, especially in Midtown and Downtown Manhattan (red and orange dots in Figure\ref{fig:diff}a). For ride-hailing, the prediction performance for most ride-hailing zones in different parts of Manhattan have been significantly improved (red and orange blocks in Figure\ref{fig:diff}b). This demonstrates that the inter-modal relations can help improve the prediction performance of most locations.

\begin{table}
  \centering \footnotesize
  \caption{Performance comparison for demand-intensive and demand-sparse stations}
    \begin{tabular}{cccccccc}
    \toprule
    \multirow{2}{*}{Models} & \multirow{2}{*}{Stations} & \multicolumn{3}{c}{Subway} & \multicolumn{3}{c}{Ride-hailing} \\
    & & RMSE & MAE & $R^2$ & RMSE & MAE & $R^2$\\
    \midrule
    \multirow{2}{*}{S-ST-MRGNN} & demand-intensive & 986.429 & 588.399 & 0.979 & 124.342 & 85.833 & 0.960 \\
    & demand-sparse & 566.788 & 274.856 & 0.632 & 53.329 & 39.601 & 0.951\\
    \multirow{2}{*}{ST-MRGNN} & demand-intensive & 941.463 & 535.174 & 0.981 & 121.431 & 82.592 & 0.962\\
    & demand-sparse & 450.600 & 235.952 & 0.756 & 49.038 & 34.369 & 0.955\\
    \bottomrule
    \end{tabular}%
  \label{table:demand_intensive_sparse}%
\end{table}%

\subsection{Spatiotemporal Analysis of Cross-Mode Dependencies} \label{res:sp analysis}
In this section, we aim to further explore the cross-mode dependencies via the relation aggregation module, which is used to summarize the spatiotemporal dependencies of each node (station/zone) of a given mode on all other nodes across another mode. Specifically, we consider $u=2$ types of spatial dependencies, i.e., geographical proximity (distance-based) and functional similarity (semantics-based) in our model. Thus, for each node of a mode, there are 4 estimated attention values, i.e. intra-modal distance-based dependency, intra-modal semantics-based dependency, inter-modal distance-based dependencies, and inter-modal semantics-based dependencies. The 4 attention values represent the information weights that a node focuses on for demand prediction, and they sum up to 1.

\begin{figure} [ht!]
  \centering
  \includegraphics[width=0.85\textwidth]{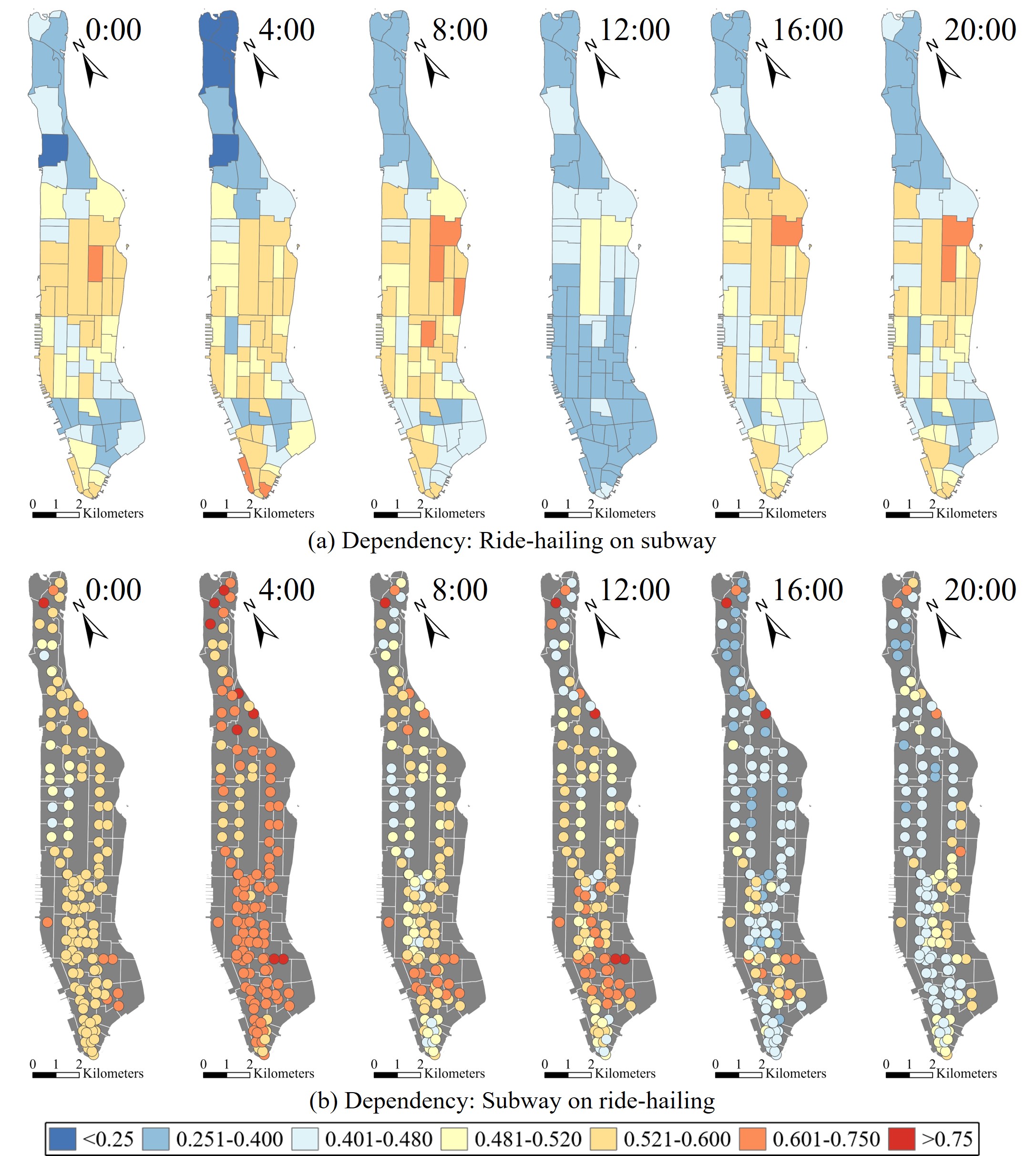}
  \caption{Spatiotemporal distribution of modal dependencies for demand prediction} \label{fig:spatial}
\end{figure}

The spatiotemporal distributions of the estimated attention weights are presented in Figure~\ref{fig:spatial}. For presentation simplicity, the distance-based and semantics-based dependencies associated with one mode are added together and meanwhile, only the inter-modal dependency is presented, as the values of intra- and inter-modal dependencies sum up to 1. The first row shows to what extent a ride-hailing zone depends on other subway stations across different time of day, and the second row is vice versa. The color represents the attention weight value. The light yellow color represents the value of inter-modal dependency near 0.5, meaning that the intra- and inter-modal dependency weights are roughly equal. Colors closer to the red indicate the station/zone depends more on inter-modal dependency, or blue for more dependency on intra-modal dependency. Based on the figure, we can find that the demand prediction for both ride-hailing and subway benefit significantly from inter-modal relations as there are considerable stations/zones with inter-modal attention weights higher than 0.5. For ride-hailing, a few zones spanning across Midtown, Uptown and Upper Manhattan benefit more from the inter-modal dependency from the afternoon rush hour (16:00) to the morning rush hour (12:00), especially the East Harlem area at the northeast Manhattan. In comparison, in the Downtown and the north part of Upper Manhattan where the ride-hailing demand are relatively more intensive and the subway demand are relatively not that high, the intra-modal dependency is higher. Correspondingly, the subway stations in these areas rely mostly on the demand of ride-hailing, rather than other subway stations, to provide information for their demand prediction. During the midnight to the start of morning rush hour (0:00 - 8:00), the inter-modal dependency is more important for nearly all subway stations. During the daytime, as the usage of subway becomes active (Figure~\ref{fig:od_t}), the weights of the inter-modal dependency decreases, and reach the lowest level at around 16:00, the typical peak hours for subway demand in Manhattan.

\begin{figure}
  \centering
  \includegraphics[width=1\textwidth]{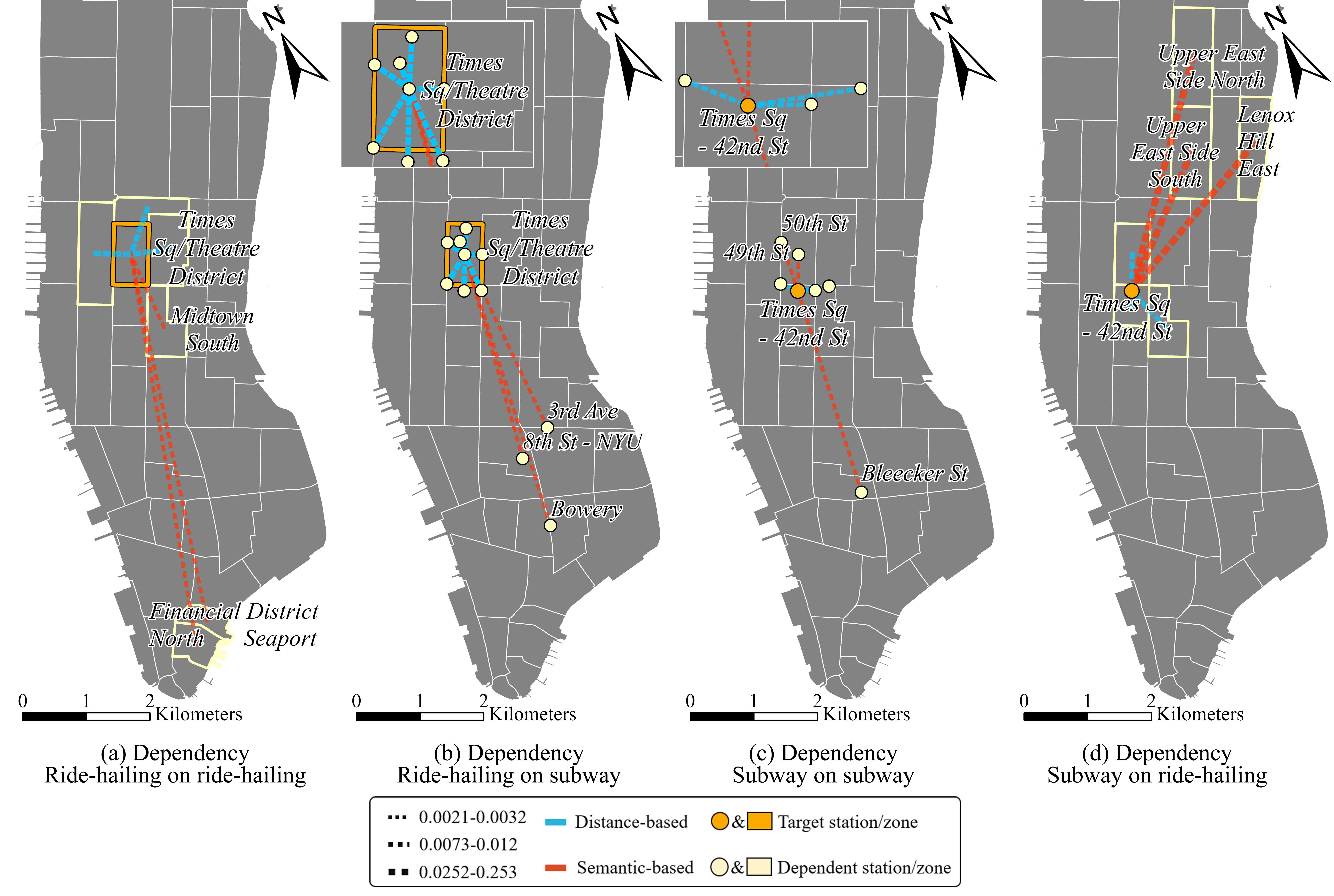}
  \caption{An example ride-hailing zone and subway station and their node-level dependencies across modes}\label{fig:time_square}
\end{figure}

We further choose the Times Square/Theatre District ride-hailing zone and the Times Square-42nd street subway station as examples and examine how their demands are dependent on other nodes in the bimodal system. Recall that the spatial dependencies between nodes for each relation is encoded with a corresponding adjacency matrix. Therefore, the node-to-node dependency for a specific relation can be calculated as a product of the relation-level attention weight and the normalized correlation weight in the adjacency matrix. Figure~\ref{fig:time_square} presents the top 3 neighborhood nodes with highest node-to-node dependency values for each spatial dependency. It is apparent that the distance-based dependencies (shown as blue dashed lines) mostly come from nearby zones/stations as expected, but semantics-based dependencies (shown as red dashed lines) can be from faraway locations. In terms of the relative importance, it seems that intra-modal distance-based dependencies are more useful than inter-modal dependencies for demand prediction related to the selected station/zone, which may be resulting from the significant spatial clustering effect of travel demand distributions. However, the importance of inter-modal semantics-based dependencies varies, as they are more important for the selected subway station, but less so for the ride-hailing zone. These examples demonstrate the usability of our proposed model to derive operational insights at the individual station/zone level.

\section{Conclusion}
This paper focuses on the problem of joint demand prediction for multimodal transportation systems, which is important for dynamically providing intelligent travel recommendations, reducing congestion, and enhancing system resilience. Compared with single-mode demand prediction, this problem is more challenging because different transportation modes may have diverse spatial units and heterogeneous spatiotemporal correlations. To address these challenges, we propose a multi-relational spatiotemporal graph neural network (ST-MRGNN) approach. Specifically, a multi-relational graph neural network (MRGNN) is developed to capture cross-mode heterogeneous spatial dependencies consisting of three steps. First, both geographic and semantic spatial dependencies between nodes associated with different modes are encoded with multiple intra- and inter-modal relation graphs. Second, a generalized graph convolution network is introduced to learn the latent representation of each node based on its neighborhood nodes through each relation. Third, we design an attention-based aggregation module to obtain the optimal combination of different relations. To further integrate correlations from spatial and temporal domains, we incorporate MRGNN layers with temporal gated convolution layers in stacked multi-relational spatiotemporal (ST-MR) blocks to jointly model heterogeneous spatiotemporal correlations. To test the model performance, extensive experiments are conducted on real-world subway and ride-hailing datasets from NYC. The results show that: (1) our proposed model is able to improve the demand prediction performance for both subway and ride-hailing, beating classic and state-of-the-art demand prediction methods; (2) the cross-mode relationships, which are encoded with multiple inter-modal relation graphs in our model, are crucial for the improvement of prediction performance; and (3) further analysis implies that the cross-mode relationships are especially helpful for demand-sparse locations. 
In addition, we visualize the spatiotemporal distributions of different relations learned from the attention mechanism and demonstrate the potential interpretability of our propose model.

In general, this study provides insights on how to adapt graph neural networks to heterogeneous mobility networks for multimodal demand prediction. The proposed model can be further improved or extended in a plethora of ways. First, when the number of relations is large, most heterogeneous graph embedding techniques, including our method, may suffer from over-parameterization. Future research can explore ways to alleviate the problem and improve the model scalability using techniques such as graph embedding \cite{vashishth2019composition}. Second, different transportation modes may have different data distributions, and simply using inter-modal relations may not be the best approach in cases where different modes exhibit distribution discrepancies. Future research can integrate domain adaptation techniques in the model to promote positive transfer across modes. Finally, although our proposed model is developed for multimodal demand prediction, it can be easily adapted for other research problems. For example, since our model shows more advantage in demand-sparse locations, as illustrated in Section~\ref{res:multi_vs_single}, it is natural to adapt our model to specifically improve the prediction performance of demand-sparse modes (e.g., bike sharing) with the help of demand-intensive modes (e.g., subway) through transfer learning.

\section*{Acknowledgements}
This research is partly supported by the Seed Fund for Basic Research for New Staff (104006019) at the University of Hong Kong.

\bibliographystyle{unsrt}  
\bibliography{main}

\begin{thebibliography}{10}

\bibitem{li2017diffusion}
Yaguang Li, Rose Yu, Cyrus Shahabi, and Yan Liu.
\newblock Diffusion convolutional recurrent neural network: Data-driven traffic
  forecasting.
\newblock {\em arXiv preprint arXiv:1707.01926}, 2017.

\bibitem{geng2019spatiotemporal}
Xu~Geng, Yaguang Li, Leye Wang, Lingyu Zhang, Qiang Yang, Jieping Ye, and Yan
  Liu.
\newblock Spatiotemporal multi-graph convolution network for ride-hailing
  demand forecasting.
\newblock In {\em Proceedings of the AAAI conference on artificial
  intelligence}, volume~33, pages 3656--3663, 2019.

\bibitem{irawan2020compete}
Muhammad~Zudhy Irawan, Prawira~Fajarindra Belgiawan, Ari Krisna~Mawira Tarigan,
  and Fajar Wijanarko.
\newblock To compete or not compete: exploring the relationships between
  motorcycle-based ride-sourcing, motorcycle taxis, and public transport in the
  jakarta metropolitan area.
\newblock {\em Transportation}, 47(5):2367--2389, 2020.

\bibitem{ye2019co}
Junchen Ye, Leilei Sun, Bowen Du, Yanjie Fu, Xinran Tong, and Hui Xiong.
\newblock Co-prediction of multiple transportation demands based on deep
  spatio-temporal neural network.
\newblock In {\em Proceedings of the 25th ACM SIGKDD International Conference
  on Knowledge Discovery \& Data Mining}, pages 305--313, 2019.

\bibitem{wang2021learning}
Qianru Wang, Bin Guo, Yi~Ouyang, Lu~Cheng, Liang Wang, Zhiwen Yu, and Huan Liu.
\newblock Learning shared mobility-aware knowledge for multiple urban travel
  demands.
\newblock {\em IEEE Internet of Things Journal}, 2021.

\bibitem{ke2021joint}
Jintao Ke, Siyuan Feng, Zheng Zhu, Hai Yang, and Jieping Ye.
\newblock Joint predictions of multi-modal ride-hailing demands: A deep
  multi-task multi-graph learning-based approach.
\newblock {\em Transportation Research Part C: Emerging Technologies},
  127:103063, 2021.

\bibitem{wang2020multi}
Senzhang Wang, Hao Miao, Hao Chen, and Zhiqiu Huang.
\newblock Multi-task adversarial spatial-temporal networks for crowd flow
  prediction.
\newblock In {\em Proceedings of the 29th ACM international conference on
  information \& knowledge management}, pages 1555--1564, 2020.

\bibitem{li2021multi}
Can Li, Lei Bai, Wei Liu, Lina Yao, and S~Travis Waller.
\newblock A multi-task memory network with knowledge adaptation for multimodal
  demand forecasting.
\newblock {\em Transportation Research Part C: Emerging Technologies},
  131:103352, 2021.

\bibitem{zhang2011seasonal}
Ning Zhang, Yunlong Zhang, and Haiting Lu.
\newblock Seasonal autoregressive integrated moving average and support vector
  machine models: prediction of short-term traffic flow on freeways.
\newblock {\em Transportation Research Record}, 2215(1):85--92, 2011.

\bibitem{antoniou2013dynamic}
Constantinos Antoniou, Haris~N Koutsopoulos, and George Yannis.
\newblock Dynamic data-driven local traffic state estimation and prediction.
\newblock {\em Transportation Research Part C: Emerging Technologies},
  34:89--107, 2013.

\bibitem{lippi2013short}
Marco Lippi, Matteo Bertini, and Paolo Frasconi.
\newblock Short-term traffic flow forecasting: An experimental comparison of
  time-series analysis and supervised learning.
\newblock {\em IEEE Transactions on Intelligent Transportation Systems},
  14(2):871--882, 2013.

\bibitem{feng2021multi}
Siyuan Feng, Jintao Ke, Hai Yang, and Jieping Ye.
\newblock A multi-task matrix factorized graph neural network for co-prediction
  of zone-based and od-based ride-hailing demand.
\newblock {\em IEEE Transactions on Intelligent Transportation Systems}, 2021.

\bibitem{moreira2013predicting}
Luis Moreira-Matias, Joao Gama, Michel Ferreira, Joao Mendes-Moreira, and Luis
  Damas.
\newblock Predicting taxi--passenger demand using streaming data.
\newblock {\em IEEE Transactions on Intelligent Transportation Systems},
  14(3):1393--1402, 2013.

\bibitem{tong2017simpler}
Yongxin Tong, Yuqiang Chen, Zimu Zhou, Lei Chen, Jie Wang, Qiang Yang, Jieping
  Ye, and Weifeng Lv.
\newblock The simpler the better: a unified approach to predicting original
  taxi demands based on large-scale online platforms.
\newblock In {\em Proceedings of the 23rd ACM SIGKDD international conference
  on knowledge discovery and data mining}, pages 1653--1662, 2017.

\bibitem{lv2014traffic}
Yisheng Lv, Yanjie Duan, Wenwen Kang, Zhengxi Li, and Fei-Yue Wang.
\newblock Traffic flow prediction with big data: a deep learning approach.
\newblock {\em IEEE Transactions on Intelligent Transportation Systems},
  16(2):865--873, 2014.

\bibitem{xu2017real}
Jun Xu, Rouhollah Rahmatizadeh, Ladislau B{\"o}l{\"o}ni, and Damla Turgut.
\newblock Real-time prediction of taxi demand using recurrent neural networks.
\newblock {\em IEEE Transactions on Intelligent Transportation Systems},
  19(8):2572--2581, 2017.

\bibitem{ke2017short}
Jintao Ke, Hongyu Zheng, Hai Yang, and Xiqun~Michael Chen.
\newblock Short-term forecasting of passenger demand under on-demand ride
  services: A spatio-temporal deep learning approach.
\newblock {\em Transportation Research Part C: Emerging Technologies},
  85:591--608, 2017.

\bibitem{yao2018deep}
Huaxiu Yao, Fei Wu, Jintao Ke, Xianfeng Tang, Yitian Jia, Siyu Lu, Pinghua
  Gong, Jieping Ye, and Zhenhui Li.
\newblock Deep multi-view spatial-temporal network for taxi demand prediction.
\newblock In {\em Proceedings of the AAAI Conference on Artificial
  Intelligence}, volume~32, 2018.

\bibitem{noursalehi2021dynamic}
Peyman Noursalehi, Haris~N Koutsopoulos, and Jinhua Zhao.
\newblock Dynamic origin-destination prediction in urban rail systems: A
  multi-resolution spatio-temporal deep learning approach.
\newblock {\em IEEE Transactions on Intelligent Transportation Systems}, 2021.

\bibitem{yu2017spatio}
Bing Yu, Haoteng Yin, and Zhanxing Zhu.
\newblock Spatio-temporal graph convolutional networks: A deep learning
  framework for traffic forecasting.
\newblock {\em arXiv preprint arXiv:1709.04875}, 2017.

\bibitem{wu2019graph}
Zonghan Wu, Shirui Pan, Guodong Long, Jing Jiang, and Chengqi Zhang.
\newblock Graph wavenet for deep spatial-temporal graph modeling.
\newblock {\em arXiv preprint arXiv:1906.00121}, 2019.

\bibitem{li2020graph}
Can Li, Lei Bai, Wei Liu, Lina Yao, and S~Travis Waller.
\newblock Graph neural network for robust public transit demand prediction.
\newblock {\em IEEE Transactions on Intelligent Transportation Systems}, 2020.

\bibitem{liang2021dynamic}
Yuebing Liang, Zhan Zhao, and Lijun Sun.
\newblock Dynamic spatiotemporal graph convolutional neural networks for
  traffic data imputation with complex missing patterns.
\newblock {\em arXiv preprint arXiv:2109.08357}, 2021.

\bibitem{zhang2019short}
Kunpeng Zhang, Zijian Liu, and Liang Zheng.
\newblock Short-term prediction of passenger demand in multi-zone level:
  Temporal convolutional neural network with multi-task learning.
\newblock {\em IEEE transactions on intelligent transportation systems},
  21(4):1480--1490, 2019.

\bibitem{zhang2020taxi}
Chizhan Zhang, Fenghua Zhu, Xiao Wang, Leilei Sun, Haina Tang, and Yisheng Lv.
\newblock Taxi demand prediction using parallel multi-task learning model.
\newblock {\em IEEE Transactions on Intelligent Transportation Systems}, 2020.

\bibitem{liu2021community}
Hao Liu, Qiyu Wu, Fuzhen Zhuang, Xinjiang Lu, Dejing Dou, and Hui Xiong.
\newblock Community-aware multi-task transportation demand prediction.
\newblock In {\em Proceedings of the AAAI Conference on Artificial
  Intelligence}, volume~35, pages 320--327, 2021.

\bibitem{jiang2021graph}
Weiwei Jiang and Jiayun Luo.
\newblock Graph neural network for traffic forecasting: A survey.
\newblock {\em arXiv preprint arXiv:2101.11174}, 2021.

\bibitem{wang2020survey}
Xiao Wang, Deyu Bo, Chuan Shi, Shaohua Fan, Yanfang Ye, and Philip~S Yu.
\newblock A survey on heterogeneous graph embedding: methods, techniques,
  applications and sources.
\newblock {\em arXiv preprint arXiv:2011.14867}, 2020.

\bibitem{schlichtkrull2018modeling}
Michael Schlichtkrull, Thomas~N Kipf, Peter Bloem, Rianne Van Den~Berg, Ivan
  Titov, and Max Welling.
\newblock Modeling relational data with graph convolutional networks.
\newblock In {\em European semantic web conference}, pages 593--607. Springer,
  2018.

\bibitem{vashishth2019composition}
Shikhar Vashishth, Soumya Sanyal, Vikram Nitin, and Partha Talukdar.
\newblock Composition-based multi-relational graph convolutional networks.
\newblock {\em arXiv preprint arXiv:1911.03082}, 2019.

\bibitem{wang2019heterogeneous}
Xiao Wang, Houye Ji, Chuan Shi, Bai Wang, Yanfang Ye, Peng Cui, and Philip~S
  Yu.
\newblock Heterogeneous graph attention network.
\newblock In {\em The World Wide Web Conference}, pages 2022--2032, 2019.

\bibitem{zhang2020relational}
Zhao Zhang, Fuzhen Zhuang, Hengshu Zhu, Zhiping Shi, Hui Xiong, and Qing He.
\newblock Relational graph neural network with hierarchical attention for
  knowledge graph completion.
\newblock In {\em Proceedings of the AAAI Conference on Artificial
  Intelligence}, volume~34, pages 9612--9619, 2020.

\bibitem{kipf2016semi}
Thomas~N Kipf and Max Welling.
\newblock Semi-supervised classification with graph convolutional networks.
\newblock {\em arXiv preprint arXiv:1609.02907}, 2016.

\bibitem{he2016deep}
Kaiming He, Xiangyu Zhang, Shaoqing Ren, and Jian Sun.
\newblock Deep residual learning for image recognition.
\newblock In {\em Proceedings of the IEEE conference on computer vision and
  pattern recognition}, pages 770--778, 2016.

\bibitem{ba2016layer}
Jimmy~Lei Ba, Jamie~Ryan Kiros, and Geoffrey~E Hinton.
\newblock Layer normalization.
\newblock {\em arXiv preprint arXiv:1607.06450}, 2016.

\bibitem{chen2015xgboost}
Tianqi Chen, Tong He, Michael Benesty, Vadim Khotilovich, Yuan Tang, Hyunsu
  Cho, et~al.
\newblock Xgboost: extreme gradient boosting.
\newblock {\em R package version 0.4-2}, 1(4):1--4, 2015.

\end{thebibliography}

\end{document}